\documentclass[twoside,leqno,twocolumn]{article}

\usepackage[letterpaper]{geometry}

\usepackage{ltexpprt}

\usepackage{graphicx}

\usepackage{comment}
\usepackage{color}

\usepackage{amsmath}

\usepackage{algorithmicx,algorithm}
\usepackage{CJK}
\usepackage{threeparttable}
\usepackage{booktabs}
\usepackage{algpseudocode}
\usepackage{graphics}
\usepackage{multirow}
\usepackage{bm}

\usepackage{subfigure}

\usepackage{bbding}
\usepackage{amssymb}

\usepackage[colorlinks]{hyperref}

\newcommand*\hh{\mathcal{H}}
\newcommand*\sss{\mathcal{S}}
\newcommand*\ttt{\mathcal{T}}
\newcommand*\rr{\mathcal{R}}

\newcommand{\bb}[1]{\bm{#1}}
\newcommand{\ms}[1]{\mathcal{#1}}

\newcommand*\red{\textcolor{red}}
\newcommand*\ie{{i.e.}}

\begin{document}

\title{\Large Contradictory Structure Learning for Semi-supervised Domain Adaptation}
\author{Can Qin\footnote{Northeastern University (\{qin.ca,yin.yu1,wang.huan\}@north- eastern.edu, wanglichenxj@gmail.com, yunfu@ece.neu.edu) }~, Lichen Wang$^*$, Qianqian Ma\footnote{Boston University (maqq@bu.edu)}~, Yu Yin$^*$, Huan Wang$^*$, Yun Fu$^*$}

\date{}

\maketitle








\begin{abstract} \small\baselineskip=9pt Current adversarial adaptation methods attempt to align the cross-domain features, whereas two challenges remain unsolved: 1) the conditional distribution mismatch and 2) the bias of the decision boundary towards the source domain. To solve these challenges, we propose a novel framework for semi-supervised domain adaptation by \textbf{unifying the learning of opposite structures (UODA)}. \textbf{UODA} consists of a generator and two classifiers (i.e., \textbf{the source-scattering classifier} and \textbf{the target-clustering classifier}), which are trained for contradictory purposes. The target-clustering classifier attempts to cluster the target features to improve intra-class density and enlarge inter-class divergence. Meanwhile, the source-scattering classifier is designed to scatter the source features to enhance the  decision boundary's smoothness. Through the alternation of source-feature expansion and target-feature clustering procedures, the target features are well-enclosed within the dilated boundary of the corresponding source features. This strategy can make the cross-domain features to be precisely aligned against the source bias simultaneously. Moreover, to overcome the model collapse through training, we progressively update the measurement of feature's distance and their representation via an adversarial training paradigm. Extensive experiments on the benchmarks of DomainNet and Office-home datasets demonstrate the superiority of our approach over the state-of-the-art methods.
\end{abstract}


\section{Introduction}
In recent years, Deep Neural Networks (DNNs) have been applied to a wide range of data mining and computer vision tasks, including recommendation system, image classification, semantic segmentation and object detection ~\cite{jing2017neural,Cordts2016Cityscapes,he2016deep,he2017mask,simonyan2014very}. Despite such great success, DNNs are eagerly hungry for enormous amounts of labeled data, which, however, is very expensive and time-consuming to collect manually.


\begin{figure*}[t]
\centering
\scalebox{1}{\includegraphics[width=0.99\textwidth]{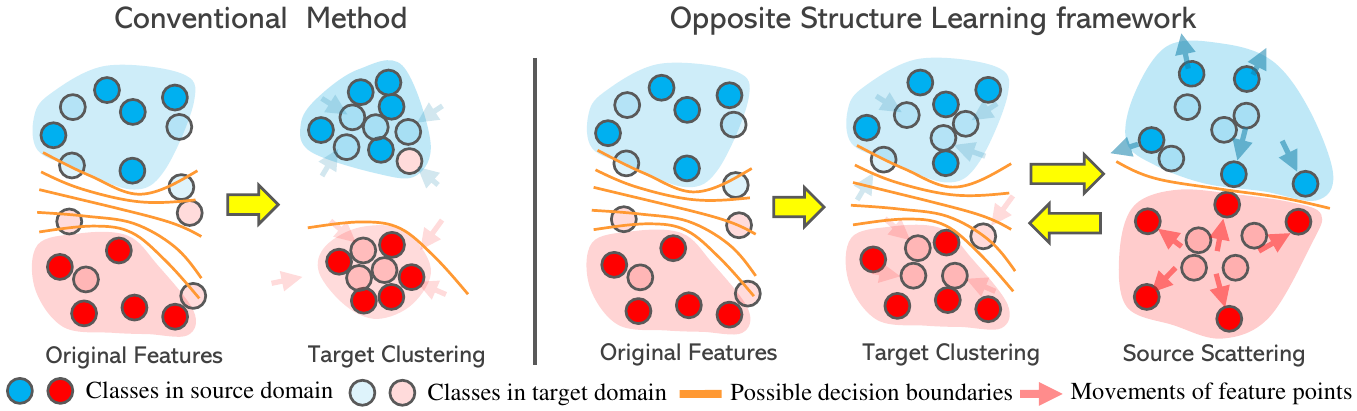}}
\caption{Illustration of the Opposite Structure Learning. Conventional method might falsely cluster features. During the alternation of \textbf{target feature clustering} and \textbf{source feature scattering}, the target features are well enclosed within the expanded boundary of corresponding source features.
} \label{f1}
\vspace{-3mm}
\end{figure*}

To solve such challenge, Domain Adaptation (DA)~\cite{peng2017visda,saito2019semi,sdm20_domain1,sdm20_domain2} has been proposed by employing the annotated data in the source domain to make it well-generalized on the label-scarce target domain. However, one crucial challenge for DA is the distribution shift (i.e., domain gap) of cross-domain features, which violates the distribution-sharing assumption of conventional machine learning algorithms. To mitigate such domain gap, feature alignment methods attempt to project the raw data into a shared space where the feature divergence or distance is minimized to learn a more universal representation. Various methods, such as Maximum Mean Discrepancy (MMD)~\cite{long2013transfer}, Correlation Alignment (CORAL)~\cite{sun2015subspace} and Geodesic Flow Kernel (GFK)~\cite{gong2012geodesic,gopalan2011domain} have been developed. Currently, adversarial domain alignment methods (i.e., DANN~\cite{ganin2016domain}, ADDA~\cite{tzeng2017adversarial}) have attracted increasing attention which utilizes a zero-sum game between a domain classifier (\ie, discriminator) and a feature generator. The features of different domains will be mixed if the discriminator cannot differentiate the source and target features. 


Recently, learning well-clustered target features proved to be helpful in conditional distributions alignment. Both DIRT-T~\cite{shu2018dirt} and MME~\cite{saito2019semi} methods applied entropy loss on target features to implicitly group them as multiple clusters in the feature space to keep the discriminative structures through adaptation. However, due to the imbalance of the labeled data between the two domains, the decision boundary is still biased towards the rich-labeling source domain, which brings two negative effects: 1) the target domain features near the boundary are easily driven to the wrong sides; 2) the boundary would cross the high-density region of target domain features to mistakenly split them.


According to our observation that the noisy labels are helpful to boost the performance on DA, we further infer that the bias of the decision boundary towards the source domain can also be mitigated by learning slightly messy and scattered source features as the regularization. To this end, a good representation for DA would be comprehensively summarized in two aspects: 1) \textbf{well-clustered target features for conditional distribution matching} and 2) \textbf{well-scattered source features to regulate the biased model}. This paper proposes a well designed semi-supervised domain adaptation framework with a generator network and two classifier networks ({i.e.,} the source-scattering classifier and the target-clustering classifier) by unifying the learning of these opposite structures.

In our approach, the target-clustering classifier attempts to group the target features in an explicit way by using conditional entropy. While the source-scattering classifier addresses the bias on the source domain by dispersing the source features in which the decision boundary becomes more smooth and lays in the middle of the separable class-wise centers. As shown in Fig.~\ref{f1}, given the few labeled target samples for support, the target features are well enclosed within the dilated boundary of the corresponding source features through the alternation of source feature expansion and target feature clustering. Moreover, the decision boundary only crosses the low-density region of target features, which further enforces the inter-class divergence. To overcome the model collapse through training, we progressively update the measurement of distance and the feature representation via an adversarial training paradigm. Our proposed model is trained in an end-to-end manner and is friendly to implementation. \footnote{The code is uploaded on \url{https://github.com/canqin001/UODA}}

In summary, our framework has three major contributions as follows: 
\vspace{-3mm}
\begin{itemize}
\item A good representation for semi-supervised domain adaptation could be empirically summarised as two contradictory aspects, including both 1) the scattered source-domain features and 2) the well-clustered target-domain features enclosed by corresponding source ones.
\vspace{-2mm}

\item Inspired by such contradictions, we propose a novel semi-supervised domain adaptation framework by unifying the learning of opposite structures across domains, which can reduce the bias of the decision boundaries towards the source domain and align the conditional distribution.
\vspace{-2mm}

\item Extensive experiments on the popular benchmarks of DomainNet~\cite{peng2019moment} and Office-home~\cite{venkateswara2017deep} demonstrate the superiority of our approach over the state-of-the-art ones.
\vspace{-2mm}
\end{itemize}


\section{Related Works}

\subsection{Unsupervised Domain Adaptation (UDA).}
There is no label accessible in the target domain in the Unsupervised Domain Adaption (UDA). The generalization of UDA algorithms can be theoretically bounded into three parts: 1) generalization on source domain, 2) the $\mathcal{H}{\vartriangle}\mathcal{H}$-divergence between the target and source features, and 3) a constant term~\cite{ben2010theory}. Due to the distribution shift of the features among different domains, the assumption, training and testing sets sharing the same distributions, of conventional machine learning methods is violated. In recent years, many UDA approaches have been proposed~\cite{cao2018unsupervised,ganin2014unsupervised,khan2016adapting,wang2018visual,qin2019pointdan,Dong_2020_CVPR,dong2020cscl} to address this problem by projecting the raw data points into a shared feature space where the cross-domain representations can be aligned by minimizing the divergence.

\begin{figure*}[t]
\centering
\scalebox{1}{\includegraphics[width=0.99\textwidth]{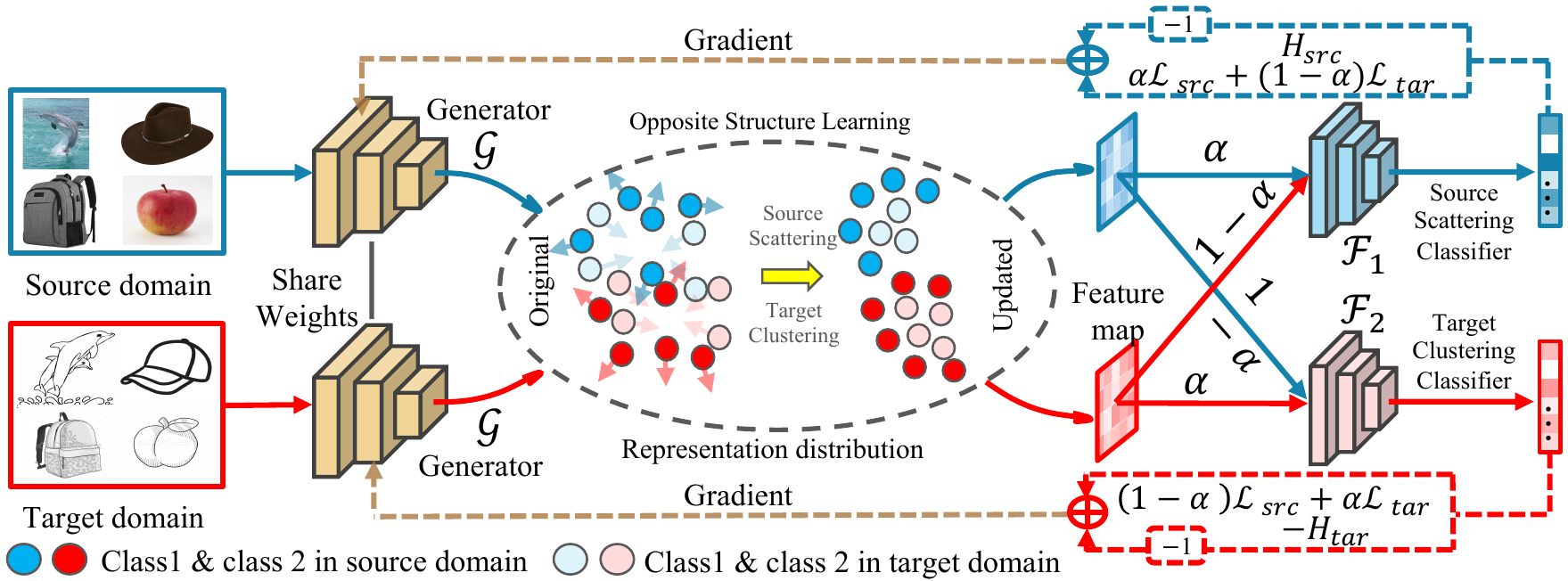}}
\vspace{-2mm}
\caption{Illustration of the proposed Unified Opposite Domain Adaptation (UODA) framework which is composed of three parts: 1) a deep feature generator network \bm{$\mathcal{G}$}, 2) a source-scattering classifier network \bm{$\mathcal{F}_1$}, and 3) a target-clustering classifier network \bm{$\mathcal{F}_2$}. \bm{$\bigoplus$} denotes element-wise sum. The gradients of entropy losses (i.e., \bm{$H_{src}$} and \bm{$-H_{tar}$}) are reversed when flowing from classifier to the generator for adversarial training. 
}\label{f2}
\vspace{-3mm}
\end{figure*}

Deep learning models are currently widely applied in UDA as the feature extraction function due to their impressive capacity in representation learning. A natural idea of DL-based UDA is to minimize certain kinds of divergence or statistical distance between the cross-domain deep features. Various methods, such as Deep Correlation Alignment (i.e., Deep-CORAL)~\cite{sun2016deep} or Deep Maximum Mean Discrepancy (i.e., Deep-MMD)~\cite{long2013transfer} have been proposed. However, the pre-defined distance is weak in measuring the distribution shift of deep features with high dimensions distributed in complicated manifolds.  Adversarial training is another popular way to learn domain-invariant representations from a generator (feature encoder) by fooling a discriminator (domain classifier) with the gradient reverse (i.e., GredRev~\cite{ganin2014unsupervised}), or GAN-based objectives (i.e., ADDA~\cite{tzeng2017adversarial}). 

Another critical issue is the conditional distributions remain mismatched after adversarial adaption. Tri-training~\cite{saito2017asymmetric} applied multiple classifiers to infer the high confident pseudo labels from different views. DIRT-T~\cite{shu2018dirt} introduced entropy loss on target features to improve the inter-class divergence and intra-class density to make decision boundaries cross the gap of clustered target features. MCD~\cite{saito2018maximum} solves this problem by replacing the discriminator with two classifiers. The decision boundaries learned by the two views provide further support to target features, which become easily aligned with source ones through adversarial training. 

\vspace{-1mm}
\subsection{Semi-supervised Domain Adaptation (SSDA).}

In Semi-supervised Domain Adaptation (SSDA), or referred as Few-shot Domain Adaptation (FSDA), the target domain samples are partially labeled. Compared with UDA, collecting limited labeled samples takes little cost, which, however, yields excellent gains in task performance. Therefore, SSDA has attracted increasing attention recently due to its high potential. In SSDA, the conditional distribution of target features can be denoted in a more detailed way due to the accessibility to labeled target samples, making the cross-domain features more precisely aligned. Although all the UDA methods can be directly applied to SSDA by adding labeled target samples to the training set, the learned model would be overfitted to the target set due to the imbalanced samples between the source and labeled target set.

To solve the problem above, \cite{tzeng2015simultaneous} firstly maximizes domain confusion by aligning the marginal distributions of cross-domain features. Then, it applies soft label scores for a matching loss to transfer the semantic relationship from the source domain to the target domain. In CCSA~\cite{motiian2017unified}, few labeled target samples are utilized to minimize the semantic alignment loss for conditional distribution matching, and it simultaneously maximizes the distance between the samples from different domains but in the same class with the help of separation loss. FADA~\cite{motiian2017few} extends CCSA by designing an adversarial training paradigm between a deep generator and multiple binary discriminators to semantically align the cross-domain features and augment the distinguishment between different classes at the same time. MME~\cite{saito2019semi} alternatively updates the estimated class-wise prototypes to maximize the entropy of the unlabeled target domain by the classifier. Furthermore, it clusters features around the estimated prototype by minimizing entropy with respect to the feature extractor. However, due to the imbalanced labeled data between source and target domains, the decision boundary is still biased towards the source domain. This situation renders the decision boundaries easily cross the density region of target-domain samples or drives the easily confused target-domain features towards the wrong side.

\section{Proposed Approach}
\vspace{-1mm}
As shown in Fig.~\ref{f2}, the framework of our model is composed of three components: 1) the generator, 2) the source-scattering classifier, and 3) the target-clustering classifier. In general, the source-scattering classifier aims to learn sparse source features while the target-clustering classifier is designed to group target features to encourage class-wise alignment. In the following subsections, we will analyze each module and explain its training procedure in details. Also, we will provide a theoretical analysis of the effectiveness of our model.

\subsection{Domain Based Classifiers.}
\vspace{-2mm}
The goal of SSDA is to learn a model which works well on the target domain based on the training set of fully labeled source-domain samples and partially labeled target-domain samples. To this end, we assume that
it is accessible to the source images ${\mathcal{S} = \{ \bb{x}^s_i, y^s_i \}^{N_s}_{i=1}}$ as well as part of the labeled target images ${\mathcal{T} = \{ \bb{x}^t_i, y^t_i \}^{N_t}_{i=1}}$, where $\bb x_i^s$, $\bb x_i^t$ represent the features, $y^s$, $y^t$ represent the corresponding labels, $N_s=|S|$, $N_t=|T|$ indicate the number of samples in $\ms S$ and $\ms T$ respectively. Moreover, the unlabeled target image set is denoted as ${\mathcal{U} = \{\bb{x}^u_i\}^{N_u}_{i=1} }$, where $N_u= |\ms U|$ and $N_u \gg N_t$. Since such two domains may have distinct marginal distributions, \textit{i.e.}, $p({\bb{x}}^s)\not=p({\bb{x}}^u)$, as well as different conditional distributions, \textit{i.e.}, $p(y^s|{\bb{x}}^s)\not=p(y^u|{\bb{x}}^u)$, the model trained only by the labeled samples usually performs poorly on the unlabeled target domain.

Our proposed method attempts to align the cross-domain features with a feature generator network $\mathcal{G}$ and two other classifier networks $\mathcal{F}_{1}(\cdot)$ and $\mathcal{F}_{2}(\cdot)$. Compared with the DA methods with only one classifier, the two classifiers are trained over different losses, hence the decision boundaries are learned from two distinct views so that they are more robust towards the noisy samples. To this end, the two classifiers are assigned with different purposes for structure optimizing. Moreover, this one-generator-two-classifiers structure is also helpful in inferring the pseudo labels with high confident from two views for the further finetuning of the model. In our approach, the generator network $\mathcal{G}(\cdot)$ is applied to extract the $d$-dimensional feature $\bb f\in\mathbb{R}^{d}$ of image $\bb{x}$ in the training set:
\begin{equation}\label{e6}
    \bb{f}=\mathcal{G} ( \bb{x}, {\Theta}_\mathcal{G}),
\end{equation}
where $\mathcal{G}(\cdot)$, parameterized by ${{\Theta}_\mathcal{G}}$, is the feature encoder function. The two classifiers $\mathcal{F}_1(\cdot)$ and $\mathcal{F}_2(\cdot)$ take the feature $\bb f$ obtained from the $\mathcal{G}$ as the input and classify them into $K$ semantical categories as following:
\begin{align}
    \bb p_1(y|\bb x)=\sigma(\ms F_1(\bb f(\bb x), {\Theta}_{\mathcal{F}_1})), \nonumber \\
    \bb p_2(y|\bb x)=\sigma(\ms F_2(\bb f(\bb x), {\Theta}_{\mathcal{F}_2})),
\end{align}
where ${\bb p_1({y}|\bb{x})}$ and ${\bb p_2({y}|\bb{x})} \in \mathbb{R}^{K}$ denote the $K$-dimensional softmax results of classification functions $\mathcal{F}_1(\cdot)$ and $\mathcal{F}_2(\cdot)$ parameterized by the ${\Theta}_{\mathcal{F}_1}$ and ${\Theta}_{\mathcal{F}_2}$ respectively.

The empirical loss is composed of two items: the task loss of source-domain $\mathcal{L}_{src}$ and the target-domain task loss $\mathcal{L}_{tar}$ formulated as:
\begin{align}
\mathcal{L}_{src} = - &\mathbb{E}_{(\bb{x}^s,y^s)  \sim \mathcal{S}}  \sum_{k=1}^{K}  \bb{1}_{[k=y^s]}{\mathrm{\log}{({p}_1({y}=y^s| \bb{x}^s)})} ,\label{e8} \\
\mathcal{L}_{tar}= - &\mathbb{E}_{(\bb{x}^t,y^t)  \sim \mathcal{T}}  \sum_{k=1}^{K}  \bb{1}_{[k=y^t]}{\mathrm{\log}{( {p}_2({y}=y^t| \bb{x}^t)})} , \label{e9}
\end{align}
where ${p}_1({y}=y^s| \bb{x}^s)$ and ${p}_2({y}=y^s| \bb{x}^s)$ represent the elements in $k$-th dimension of the softmax conditional probability $\bb p_1(y|\bb x)$ and $\bb p_2(y|\bb x)$ respectively.

To train the diversified classifiers where $\Theta_{\mathcal{F}_1} \neq \Theta_{\mathcal{F}_2}$, as illustrated in Fig~\ref{f2}, we apply asymmetric weights on both supervision losses $\mathcal{L}_{src}$ and $\mathcal{L}_{tar}$ according to a hyper-parameter $\alpha$ where the task loss for $\mathcal{F}_1(\cdot)$ is $\alpha \mathcal{L}_{src} + (1- \alpha)\mathcal{L}_{tar} $, and $(1- \alpha) \mathcal{L}_{src} +  \alpha \mathcal{L}_{tar} $ for $\mathcal{F}_2(\cdot)$ to enforce the divergence of two classifiers. The more detailed sensitive analysis of $\alpha$ would be described later in the Sec.~\ref{s4.4}.

\subsection{Opposite Structure Learning.}
Apart from learning the basic decision boundaries relying on the supervision losses, we propose the Unified Opposite Structure Learning (UOSA) which employs two classifiers to learn well-scattered source features and well-clustered target features at the same time.

In Dirt-T~\cite{shu2018dirt} and MME~\cite{saito2019semi}, it has been proved that the minimization of conditional entropy on predicted softmax scores is an effective way to cluster the features by enforcing the high-confident predictions. In this way, the features would be gathered and drove apart from the decision boundary. Followed by these, we apply the conditional entropy loss of unannotated target samples to the target-clustering classifier $\mathcal{F}_2(\cdot)$ to learn the well-clustered target features:
\begin{equation}
   H_{tar}= - \mathbb{E}_{\bb{x}^u  \sim \mathcal{U}} \sum_{k=1}^{K}  [{p_2}({y}=k| \bb{x}^u)
    {\mathrm{\log} \ {{p_2}({y}=k| \bb{x}^u)}}] , \label{e10}
\end{equation}
where ${p_2}({y}=k| \bb{x}^u)$ denotes the possibility of data $\bb{x}^u$ as the class $k$ which is the element at the $k$-th dimension of softmax score vector $\bb p_2(y|\bb x^u)=\sigma(\ms F_2(\bb f(\bb x^u)))$. Since the labeled target samples in $\mathcal{T}$ only cover a tiny part compared with those in $\mathcal{U}$, there is no need to involve them for contradict structure optimization.

Feature scattering can be regarded as the inverse process of feature clustering. To this end, we will also take conditional entropy loss $H_{src}$ to implement the source feature expansion. The definition of $H_{src}$ is given as following:
\begin{equation}
   H_{src}= -\mathbb{E}_{\bb{x}^s  \sim \mathcal{S}} \sum_{k=1}^{K}  [{p_1}({y}=k| \bb{x}^s)
    {\mathrm{\log} \ {{ p_1}({y}=k| \bb{x}^s)}}] , \label{e11}
\end{equation}
where ${p_1}({y}=k| \bb{x}^s)$ denotes the possibility of the source image $\bb{x}^s$ as the class $k$ which is the element at the $k$-th dimension of vector $\bb p_1(y|\bb x^s)=\sigma(\ms F_1(\bb f(\bb x^s)))$.

\subsection{Training Procedure.}
The training procedure consists of the optimization of 3 groups of parameters, \textit{i.e, }${\Theta}_{\mathcal{F}_{1}}$, ${\Theta}_{\mathcal{F}_{2}}$ and ${\Theta}_\mathcal{G}$.  To optimize the classifiers, we firstly apply the asymmetric weights on the task losses to minimize the empirical risk and enforce the difference between ${\Theta}_{\mathcal{F}_{1}}$ and ${\Theta}_{\mathcal{F}_{2}}$. Then, we maximize the target-domain entropy loss on $\mathcal{F}_2(\cdot)$ to update the class-wise prototypes and minimize the source-domain entropy loss on $\mathcal{F}_1(\cdot)$ to make source features slightly gathered:
\begin{align}
    {\Theta}_{\mathcal{F}_{1}}^{*} =&\mathop {\arg \min} \limits_{{\Theta}_{\mathcal{F}_{1}}} \alpha \mathcal{L}_{src} + (1- \alpha)\mathcal{L}_{tar} + \beta H_{src},\label{F_1}\\
    {\Theta}_{\mathcal{F}_{2}}^{*} =&\mathop {\arg \min} \limits_{{\Theta}_{\mathcal{F}_{2}}} (1- \alpha) \mathcal{L}_{src} +  \alpha \mathcal{L}_{tar} - \lambda H_{tar}, \label{F_2}
\end{align}
where $\lambda$ and $\beta$ are hyper-parameters used to balance the influence of conditional entropy loss and supervision loss. The minimization of conditional entropy loss enforces the confidence of classification results by driving the features away from the decision boundary which implicitly makes the features gathered. In turn, maximizing the conditional entropy loss leads to the feature scattering. 

To progressively cluster target features and disperse source features, the generator is optimized by the reversal of entropy loss where we minimize the target entropy loss for grouping and maximize the source entropy loss for scattering. The task losses are summed for empirical risk minimization:
\begin{align}\label{G}
{\Theta}_\mathcal{G}^{*} =&\mathop {\arg \min} \limits_{{\Theta}_\mathcal{G}} \mathcal{L}_{src} +  \mathcal{L}_{tar} - \beta H_{src} + \lambda H_{tar}.
\end{align}

The whole framework is trained in an end-to-end manner with the help of gradient reversal~\cite{ganin2016domain} for adversarial training and would continue to loop until reaching the certain epochs. More details of our training algorithm are shown in the \textbf{Algorithm 1}.

\begin{algorithm}[t]\label{al1}
\caption{Unified Opposite Structure Learning for Semi-supervised Domain Adaptation (UODA)}
{\bf Input:} \\
Labeled source set $\mathcal{S}$, labeled target set $\mathcal{T}$ and unlabeled target set $\mathcal{U}$. The number of training epochs \({T}\).  The hyper-parameters $\alpha$, $\beta$ and $\lambda$.\\
{\bf Output:}\\
The final parameters ${\Theta}_\mathcal{G}^{*}$, $ {\Theta}_{\mathcal{F}_1}^{*}$ and $ {\Theta}_{\mathcal{F}_2}^{*}$
\begin{algorithmic}[1]

\State Randomly initialize the parameters  ${\Theta}_\mathcal{G}^{0}$, ${\Theta}_{\mathcal{F}_1}^{0}$, ${\Theta}_{\mathcal{F}_2}^{0}$ of $\mathcal{G}$, ${\mathcal{F}_1}$ and ${\mathcal{F}_2}$.

\State \(t \leftarrow 0\)

\While{\(t<T\)}
    \State \(t \leftarrow t + 1\).
    \State update $ {\Theta}_{\mathcal{F}_1}^{t-1}$ to  $ {\Theta}_{\mathcal{F}_1}^{t}$ by Eq.~(\ref{F_1}).
    \State update $ {\Theta}_{\mathcal{F}_2}^{t-1}$ to  $ {\Theta}_{\mathcal{F}_2}^{t}$ by Eq.~(\ref{F_2}).
    \State update ${\Theta}_\mathcal{G}^{t-1}$ to ${\Theta}_\mathcal{G}^{t}$ by Eq.~(\ref{G}).
\EndWhile
\State \Return ${\Theta}_\mathcal{G}^{*}$ $\leftarrow$ ${\Theta}_\mathcal{G}^{T}$, 
$ {\Theta}_{\mathcal{F}_1}^{*}$ $\leftarrow$
$ {\Theta}_{\mathcal{F}_1}^{T}$ and 
$ {\Theta}_{\mathcal{F}_2}^{*}$ $\leftarrow$ 
$ {\Theta}_{\mathcal{F}_2}^{T}$.
\end{algorithmic}
\end{algorithm}

\section{Experiments}

\subsection{Experiments Setup.}

\textbf{Implementation Details.} We deploy the ResNet34~\cite{he2016deep} and VGG16~\cite{simonyan2014very} as the backbones of the generator $\mathcal{G}(\cdot)$. The two classifiers $\mathcal{F}_1(\cdot)$ and $\mathcal{F}_2(\cdot)$ take a two-layer MLP with randomly initialed weights. To optimize the proposed model, we take the momentum Stochastic Gradient Descent (SGD) as the optimizer on PyTorch~\cite{NEURIPS2019_9015}. The learning rate is assigned as $0.01$, and the momentum is $0.9$ with weight decay $0.0005$. The hyper-parameters $\alpha$, $\beta$ and $\lambda$ on losses are assigned as $0.75$, $0.1$, and $0.1$ respectively. We regard the parts of high-confident unlabeled target samples as the pseudo labels. 


\begin{table*}[t]
\begin{center}
\caption{ {Quantitative results (\%) on the DomainNet~\cite{peng2019moment} under ResNet-34~\cite{he2016deep} and VGG-16~\cite{simonyan2014very}. }}
\vspace{1mm}
\label{t1}
\scalebox{0.82}{
\begin{threeparttable}
 \centering
  \begin{tabular}{|c|cccccccccccccc|cc|}
   \hline \hline
\multicolumn{17}{|c|}{ResNet-34}  \\\hline
   \multirow{2}{*}{Methods} & \multicolumn{2}{c}{R$\rightarrow$C} & \multicolumn{2}{c}{R$\rightarrow$P} & \multicolumn{2}{c}{P$\rightarrow$C} & \multicolumn{2}{c}{C$\rightarrow$S} & \multicolumn{2}{c}{S$\rightarrow$P} & \multicolumn{2}{c}{R$\rightarrow$S} & \multicolumn{2}{c}{P$\rightarrow$R} & \multicolumn{2}{|c|}{Avg}\\
   
   &{1$_{shot}$} &{3$_{shot}$} &{1$_{shot}$} &{3$_{shot}$} &{1$_{shot}$} &{3$_{shot}$} &{1$_{shot}$} &{3$_{shot}$} &{1$_{shot}$} &{3$_{shot}$} &{1$_{shot}$} &{\small{3-shot}} &{1$_{shot}$} &{3$_{shot}$} &{1$_{shot}$} &\multicolumn{1}{c|}{3$_{shot}$} \\
   \hline
   
\multicolumn{1}{|c|}{S+T} &55.6&60.0  &60.6&62.2   &56.8&59.4  &50.8&55.0  &56.0&59.5  &46.3&50.1 &71.8&73.9 &56.9&\multicolumn{1}{c|}{60.0}  \\

\multicolumn{1}{|c|}{DANN~\cite{ganin2016domain}} &58.2&59.8  &61.4&62.8   &56.3&59.6  &52.8&55.4  &57.4&59.9  &52.2&54.9 &70.3&72.2 &58.4&\multicolumn{1}{c|}{60.7}  \\

\multicolumn{1}{|c|}{ADR~\cite{saito2017adversarial}} &57.1&60.7  &61.3&61.9   &57.0&60.7  &51.0&54.4  &56.0&59.9  &49.0&51.1 &72.0&74.2 &57.6&\multicolumn{1}{c|}{60.4} \\

 \multicolumn{1}{|c|}{CDAN~\cite{long2018conditional}} &65.0&69.0  &64.9&67.3   &63.7&68.4  &53.1&57.8  &63.4&65.3  &54.5&59.0 &73.2&78.5 &62.5&\multicolumn{1}{c|}{66.5} \\
 
  \multicolumn{1}{|c|}{ENT~\cite{grandvalet2005semi}} &65.2&71.0  &65.9&69.2   &65.4&71.1  &54.6&60.0  &59.7&62.1  &52.1&61.1 &75.0&78.6 &62.6&\multicolumn{1}{c|}{67.6} \\
 
\multicolumn{1}{|c|}{MME~\cite{saito2019semi}} &70.0&72.2  &67.7&69.7   &69.0&71.7  &56.3&61.8  &64.8&66.8  &61.0&61.9 &76.1&78.5 &66.4&\multicolumn{1}{c|}{68.9} \\

\multicolumn{1}{|c|}{BNM~\cite{cui2020towards}} &66.8&68.7  &67.3&68.6   &66.7&69.3  &58.2&58.3  &63.9&65.6  &59.1&60.5 &76.4&78.1 &65.5&\multicolumn{1}{c|}{67.0} \\

\multicolumn{1}{|c|}{Ours} &\textbf{72.7}&\textbf{75.4}  &\textbf{70.3}&\textbf{71.5} &\textbf{69.8}&\textbf{73.2}  &\textbf{60.5}&\textbf{64.1} &\textbf{66.4}&\textbf{69.4}  &\textbf{62.7}&\textbf{64.2} &\textbf{77.3}&\textbf{80.8}  &\textbf{68.5}&\multicolumn{1}{c|}{\textbf{71.2}} \\

\hline 
\multicolumn{17}{|c|}{VGG-16}  \\\hline

   
   
\multicolumn{1}{|c|}{S+T} &49.0&52.3  &55.4&56.7   &47.7&51.0  &43.9&48.5  &50.8&55.1  &37.9&45.0 &69.0&71.7 &50.5&\multicolumn{1}{c|}{54.3}\\

\multicolumn{1}{|c|}{DANN~\cite{ganin2016domain}} &43.9&56.8  &42.0&57.5   &37.3&49.2  &46.7&48.2  &51.9&55.6  &30.2&45.6 &65.8&70.1 &45.4&\multicolumn{1}{c|}{54.7}\\

\multicolumn{1}{|c|}{ADR~\cite{saito2017adversarial}} &48.3&50.2  &54.6&56.1   &47.3&51.5  &44.0&49.0  &50.7&53.5  &38.6&44.7 &67.6&70.9 &50.2&\multicolumn{1}{c|}{53.7} \\

 \multicolumn{1}{|c|}{CDAN~\cite{long2018conditional}} &57.8&58.1  &57.8&59.1   &51.0&57.4  &42.5&47.2  &51.2&54.5  &42.6&49.3 &71.7&74.6 &53.5&\multicolumn{1}{c|}{57.2} \\
 
 \multicolumn{1}{|c|}{ENT~\cite{grandvalet2005semi}} &39.6&50.3  &43.9&54.6   &26.4&47.4  &27.0&41.9  &29.1&51.0  &19.3&39.7 &68.2&72.5 &36.2&\multicolumn{1}{c|}{51.1} \\

\multicolumn{1}{|c|}{MME~\cite{saito2019semi}} &60.6&64.1  &63.3&63.5   &57.0&60.7  &50.9&55.4  &\textbf{60.5}&60.9  &\textbf{50.2}&54.8 &72.2&75.3 &59.2&\multicolumn{1}{c|}{62.1} \\

\multicolumn{1}{|c|}{BNM~\cite{cui2020towards}} &57.3&60.3  &59.1&60.3   &54.6&59.1  &48.6&53.3  &56.1&58.4  &44.1&50.2 &71.1&73.8 &55.9&\multicolumn{1}{c|}{59.3} \\

\multicolumn{1}{|c|}{Ours} &\textbf{62.2}&\textbf{66.2}  &\textbf{63.6}&\textbf{65.7} &\textbf{59.4}&\textbf{65.1}  &\textbf{52.3}&\textbf{57.6} &59.2&\textbf{63.2}  &49.6&\textbf{55.9} &\textbf{74.1}&\textbf{76.3}  &\textbf{60.1}&\multicolumn{1}{c|}{\textbf{64.3}} \\

 \hline \hline 
\end{tabular}
\renewcommand{\labelitemi}{}
\end{threeparttable}
}
\vspace{-7mm}
\end{center}
\end{table*}

\textbf{Benchmarks.}  The baseline methods and our proposed approach are evaluated on the latest DA benchmarks including DomainNet~\cite{peng2019moment} and Office-home~\cite{venkateswara2017deep}. For a fair evaluation, we take the same protocol of MME~\cite{saito2019semi} where $4$ domains including \textit{Real} (\textbf{R}), \textit{Painting} (\textbf{P}), \textit{Clipart} (\textbf{C})  and \textit{Sketch} (\textbf{S}) with $126$ classes from DomainNet are picked for evaluation. There are 7 adaptation scenarios organized by these 4 domains which yield different scales of domain gap. The Office-home dataset consists of 4 domains including \textit{Real} (\textbf{R}), \textit{Clipart} (\textbf{C}), \textit{Art} (\textbf{A}) and \textit{Product} (\textbf{P}) with $65$ classes. The quantity of samples in Office-home is less than those of DomainNet. To fully evaluate UODA, we have organized 12 adaptation scenarios in total to evaluate the proposed method.

%

\textbf{Evaluation.} Following the protocol of ~\cite{saito2019semi}, given the labeled samples in both domains and other unlabeled target samples for training, all the models will be evaluated on the unseen target samples (i.e., test set). We repeat the experiments of our proposed method 3 times to report the average top-1 classification accuracy in all tables. All the methods will be evaluated under the one-shot and three-shot settings where there are one or three labeled samples per class in the target domain.

\textbf{Baselines.} \textbf{S+T} is only trained by labeled source and target images without adaptation. \textbf{DANN}~\cite{ganin2016domain} is an adversarial adaptation method that applies a discriminator to confuse the source and target features. \textbf{ADR}~\cite{saito2017adversarial} is a GAN-based method applied to learn both domain-invariant and discriminative features. \textbf{CDAN}~\cite{long2018conditional} is designed for UDA which employs the entropy to control the uncertainty of predictions results to enforce the transferability. \textbf{ENT}~\cite{grandvalet2005semi} is a non-adversarial method which minimizes both the empirical risk and the conditional entropy on unlabeled targets samples. \textbf{MME}~\cite{saito2019semi} is the state-of-the art SSDA method where the conditional entropy of the unlabeled target samples is minimized by the adversarial training. \textbf{BNM}~\cite{cui2020towards} is the latest UDA method which attempts to maximize nuclear-norm to learn both transferable and discriminative representations of cross-domain features.

\subsection{Results on DomainNet.}
The quantitative results on DomainNet dataset are summarized in Table~\ref{t1}. It is easily observed that the proposed method outperforms the baselines on most of adaptation scenarios. Although the largest domain gap appears on the adaptation scenario \textit{Real} to \textit{Sketch}, ours exhibits its superiority in aligning the features of different domains based on ResNet34. On the easiest adaptation scenario \textit{Painting} to \textit{Real}, the accuracy of ours on 3-shot is over 80\% which indicates its great potential to put into practice. Compare the performance of both tasks, the improvements on the 1-shot SSDA is slightly inferior to those of the 3-shot on both backbones. This is because more labeled target examples are helpful to indicate the conditional distributions of target domain features which contributes a lot to the discriminative feature alignment. Moreover, our framework also relies on the representation learning ability of backbones where the performance gain is larger on ResNet34 than those of VGG16.

\begin{table*}[t]
\begin{center}
 \caption{{Quantitative results (\%) on Office-home~\cite{venkateswara2017deep} under the backbone of VGG-16~\cite{simonyan2014very}.} }
\vspace{1mm}
\label{t2}
\scalebox{0.88}{
\begin{threeparttable}
 \centering
  \begin{tabular}{|cccccccccccccc|}
   \hline
\multicolumn{14}{|c|}{ONE-SHOT}  \\\hline
   \multicolumn{1}{|c|}{Methods} & {R$\rightarrow$C} & {R$\rightarrow$P} & {R$\rightarrow$A} & {P$\rightarrow$R} & {P$\rightarrow$C} & {P$\rightarrow$A} & {A$\rightarrow$P} & {A$\rightarrow$C} & {A$\rightarrow$R} & {C$\rightarrow$R} & {C$\rightarrow$A} & {C$\rightarrow$P}  & \multicolumn{1}{|c|}{{Avg}}\\\hline
\multicolumn{1}{|c|}{S+T} &39.5 &75.3 &61.2 &71.6 &37.0 &52.0 &63.6 &37.5 &69.5 &64.5 &51.4  &65.9 &\multicolumn{1}{|c|}{57.4}  \\
\multicolumn{1}{|c|}{DANN~\cite{ganin2016domain}} &\textbf{52.0} &75.7 &62.7 &72.7 &45.9 &51.3 &64.3 &44.4 &68.9 &64.2 &52.3  &65.3 &\multicolumn{1}{|c|}{60.0}  \\
\multicolumn{1}{|c|}{ADR~\cite{saito2017adversarial}} &39.7 &76.2 &60.2 &71.8 &37.2 &51.4 &63.9 &39.0 &68.7 &64.8 &50.0  &65.2 &\multicolumn{1}{|c|}{57.4} \\
 \multicolumn{1}{|c|}{CDAN~\cite{long2018conditional}} &43.3 &75.7 &60.9 &69.6 &37.4 &44.5 &67.7 &39.8 &64.8 &58.7  &41.6 &66.2 &\multicolumn{1}{|c|}{55.8} \\
  \multicolumn{1}{|c|}{ENT~\cite{grandvalet2005semi}} &23.7 &77.5 &64.0 &74.6 &21.3 &44.6 &66.0 &22.4 &70.6 &62.1  &25.1 &67.7 &\multicolumn{1}{|c|}{51.6} \\
 
\multicolumn{1}{|c|}{MME~\cite{saito2019semi}} &49.1 &78.7 &65.1 &74.4 &\textbf{46.2} &56.0 &68.6 &\textbf{45.8} &72.2 &68.0 &57.5  &71.3 &\multicolumn{1}{|c|}{62.7} \\

\multicolumn{1}{|c|}{BNM~\cite{cui2020towards}} &51.0 &79.5 &62.8 &72.3 &44.0 &51.8 &67.1 &45.7 &68.4 &65.3 &52.7  &69.1 &\multicolumn{1}{|c|}{60.8} \\

\multicolumn{1}{|c|}{Ours} &49.6 &\textbf{79.8}  &\textbf{66.1} &\textbf{75.4} &{45.5} &\textbf{58.8} &\textbf{72.5} &{43.3} &\textbf{73.3} &\textbf{70.5} &\textbf{59.3} &\textbf{72.1}  &\multicolumn{1}{|c|}{\textbf{63.9}} \\
     \hline 
     
 \multicolumn{14}{|c|}{THREE-SHOT}  \\\hline
\multicolumn{1}{|c|}{S+T} &49.6 &78.6 &63.6 &72.7 &47.2 &55.9 &69.4 &47.5 &73.4 &69.7 &56.2  &70.4 &\multicolumn{1}{|c|}{62.9}  \\
\multicolumn{1}{|c|}{DANN~\cite{ganin2016domain}} &56.1 &77.9 &63.7 &73.6 &52.4 &56.3 &69.5 &50.0 &72.3 &68.7 &56.4  &69.8 &\multicolumn{1}{|c|}{63.9}  \\
\multicolumn{1}{|c|}{ADR~\cite{saito2017adversarial}} &49.0 &78.1 &62.8 &73.6 &47.8 &55.8 &69.9 &49.3 &73.3 &69.3 &56.3  &71.4 &\multicolumn{1}{|c|}{63.0} \\
 \multicolumn{1}{|c|}{CDAN~\cite{long2018conditional}} &50.2 &80.9 &62.1 &70.8 &45.1 &50.3 &74.7 &46.0 &71.4 &65.9  &52.9 &71.2 &\multicolumn{1}{|c|}{61.8} \\
  \multicolumn{1}{|c|}{ENT~\cite{grandvalet2005semi}} &48.3 &81.6 &65.5 &76.6 &46.8 &56.9 &73.0 &44.8 &75.3 &72.9  &59.1 &77.0 &\multicolumn{1}{|c|}{64.8} \\
 
\multicolumn{1}{|c|}{MME~\cite{saito2019semi}} &56.9 &82.9 &65.7 &76.7 &53.6 &59.2 &75.7 &54.9 &75.3 &72.9 &61.1  &76.3 &\multicolumn{1}{|c|}{67.6} \\

\multicolumn{1}{|c|}{BNM~\cite{cui2020towards}} &56.0 &81.2 &64.7 &73.8 &52.7 &55.8 &72.1 &51.7 &73.2 &70.3 &57.0  &73.3 &\multicolumn{1}{|c|}{65.2} \\

\multicolumn{1}{|c|}{Ours} &\textbf{57.6} &\textbf{83.6}  &\textbf{67.5} &\textbf{77.7} &\textbf{54.9} &\textbf{61.0} &\textbf{77.7} &\textbf{55.4} &\textbf{76.7} &\textbf{73.8} &\textbf{61.9} &\textbf{78.4}  &\multicolumn{1}{|c|}{\textbf{68.9}} \\
\hline
\end{tabular}
\end{threeparttable}
}
\end{center}
\vspace{-1mm}
\end{table*}

\begin{figure*}[t]
\centering
\scalebox{1}{\includegraphics[width=1.0\textwidth]{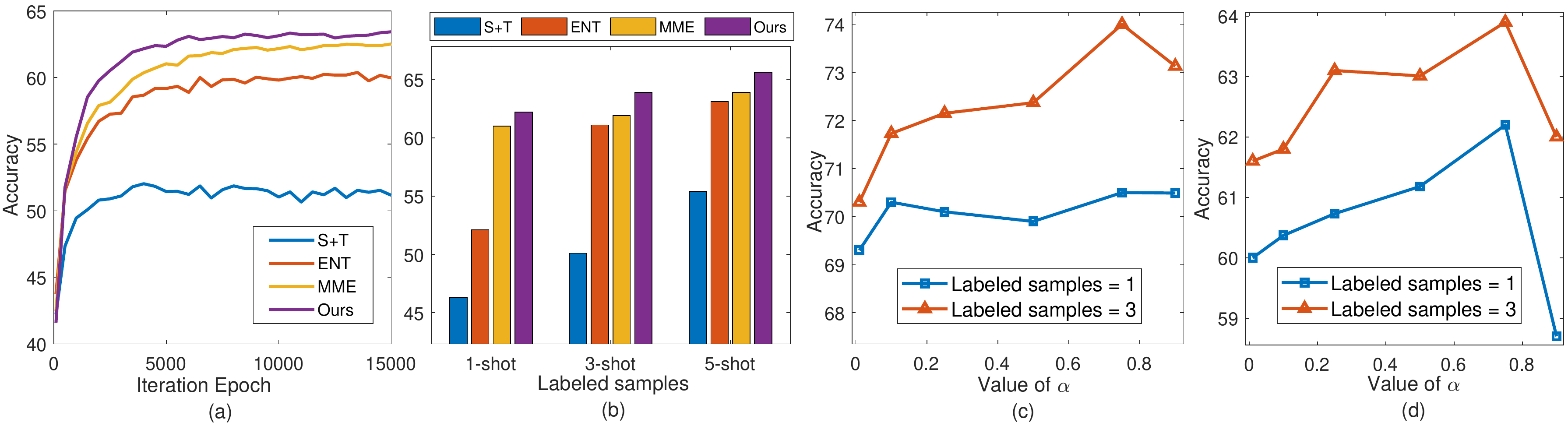}}
\vspace{-3mm}
\caption{(a) Convergence analysis on \textbf{R$\rightarrow$S}. (b) Comparison under 1-shot, 3-shot and 5-shot settings on \textbf{R$\rightarrow$S}. (c) Performance of UODA in different value of $\alpha$ on \textbf{R$\rightarrow$C} and (d) \textbf{R$\rightarrow$S}.
} \label{f5}
\vspace{-2mm}
\end{figure*}

\subsection{Results on Office-Home.}
The quantitative results and comparison on the benchmark Office-home are summarized in Table~\ref{t2}. It can be seen that the baseline methods are outperformed by our proposed method on most of adaptation scenarios which comprehensively demonstrates the superiority of our proposed method, especially on the task of three-shot SSDA. However, on the most challenging adaptation scenarios \textit{Painting} to \textit{Clipart} (i.e., \textbf{P$\rightarrow$C}) and \textit{Art} to \textit{Clipart} (i.e., \textbf{A$\rightarrow$C}) under the setting of one-shot SSDA, ours is slightly defeated by MME approach. It indicates that the maximization of source entropy has the chance to introduce some negative transferring to feature alignment. This phenomenon might be explained as the scattering of source features can drive the noisy target features to the wrong side if there are no strong constraints like enough labeled target samples or other kinds of prior knowledge to regulate. In this way, we further conclude that a limited number target labels are necessary.

\subsection{Analyses.}\label{s4.4}
\begin{figure*}[t]
    \centering
    \subfigure[]{
    \centering
    \includegraphics[width=1.25in,height=1.0in]{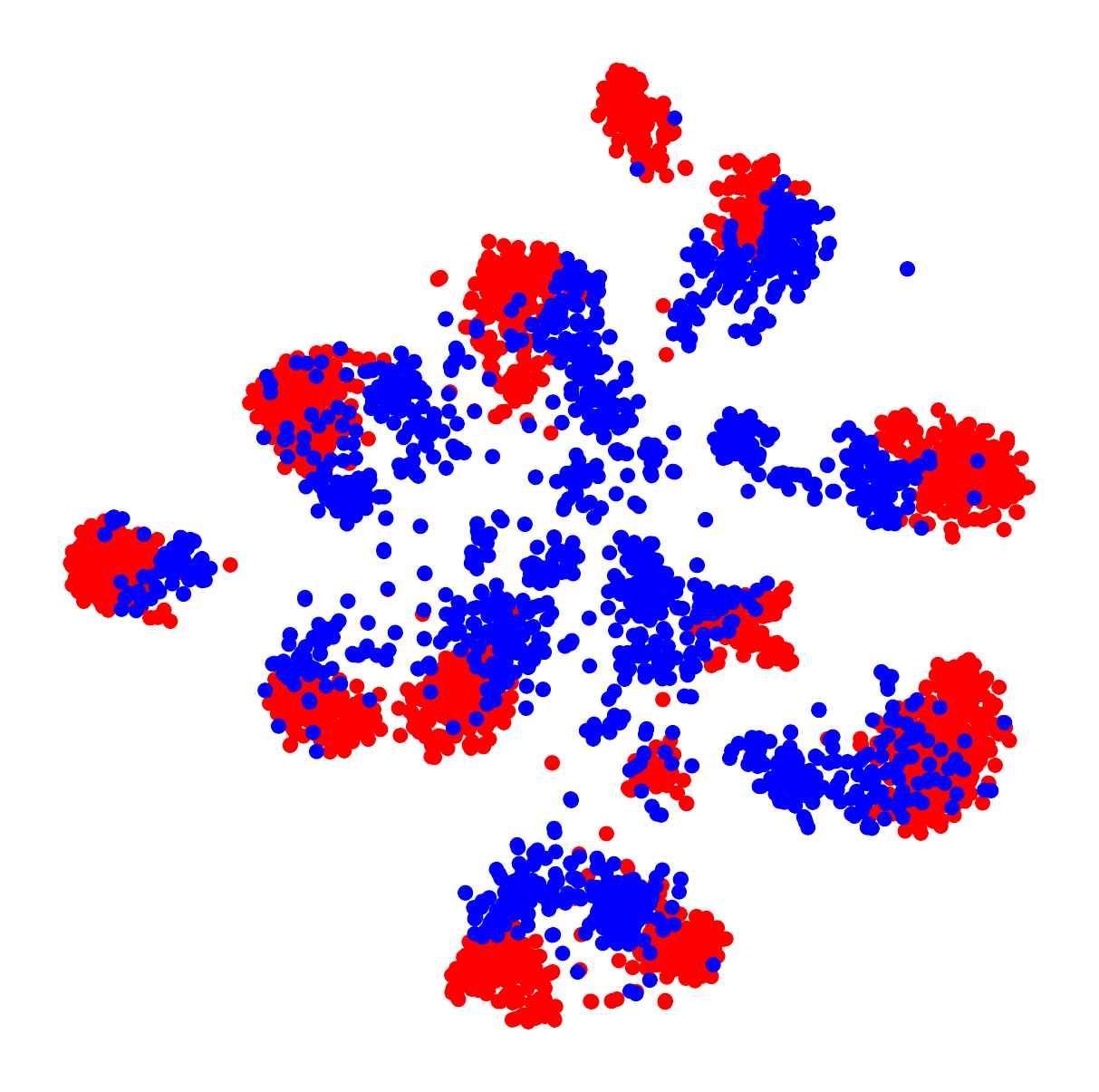}
    }
    \subfigure[]{
    \centering
    \includegraphics[width=1.25in,height=1.0in]{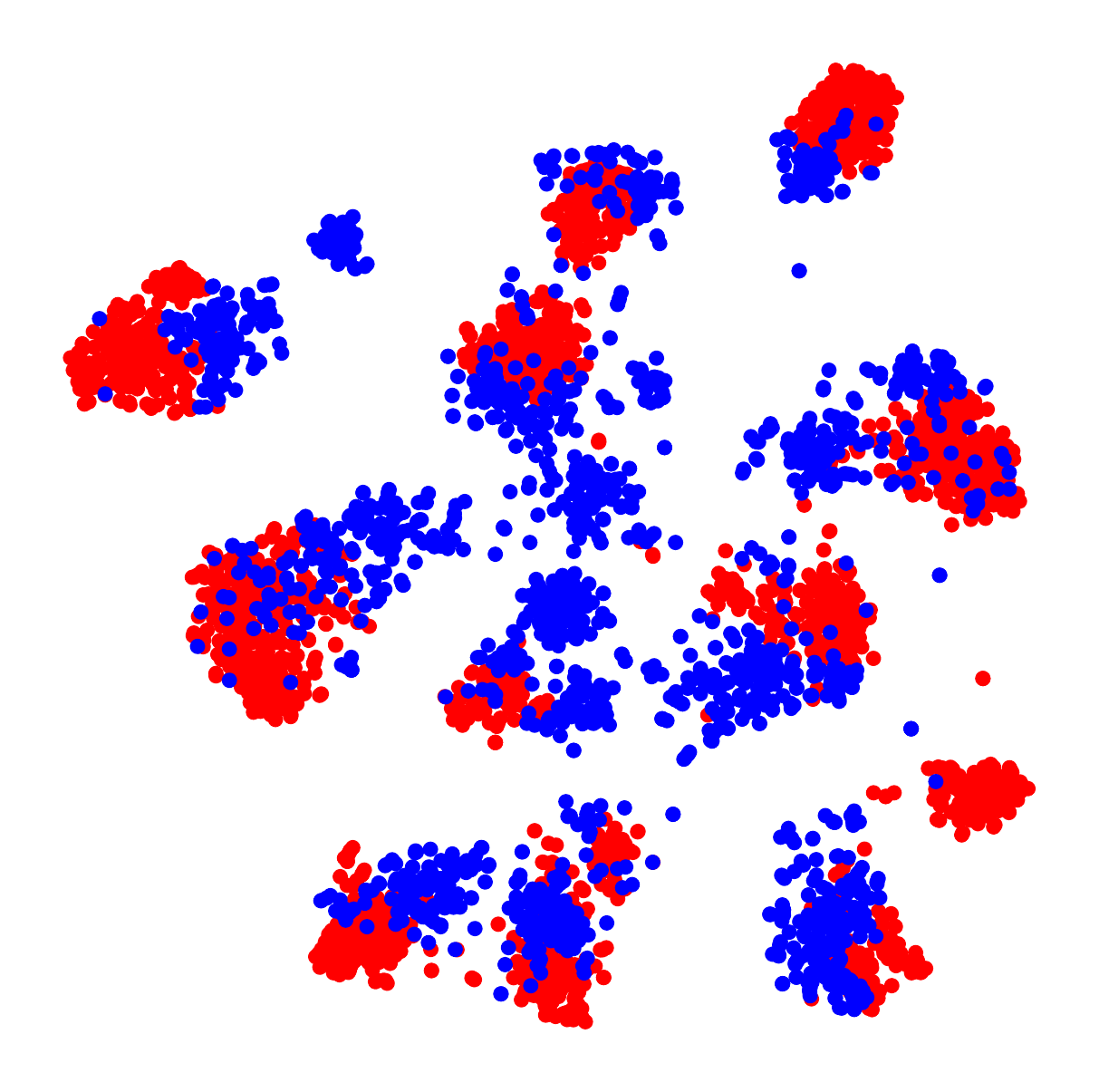}
    }
    \subfigure[]{
    \centering
    \includegraphics[width=1.25in,height=1.0in]{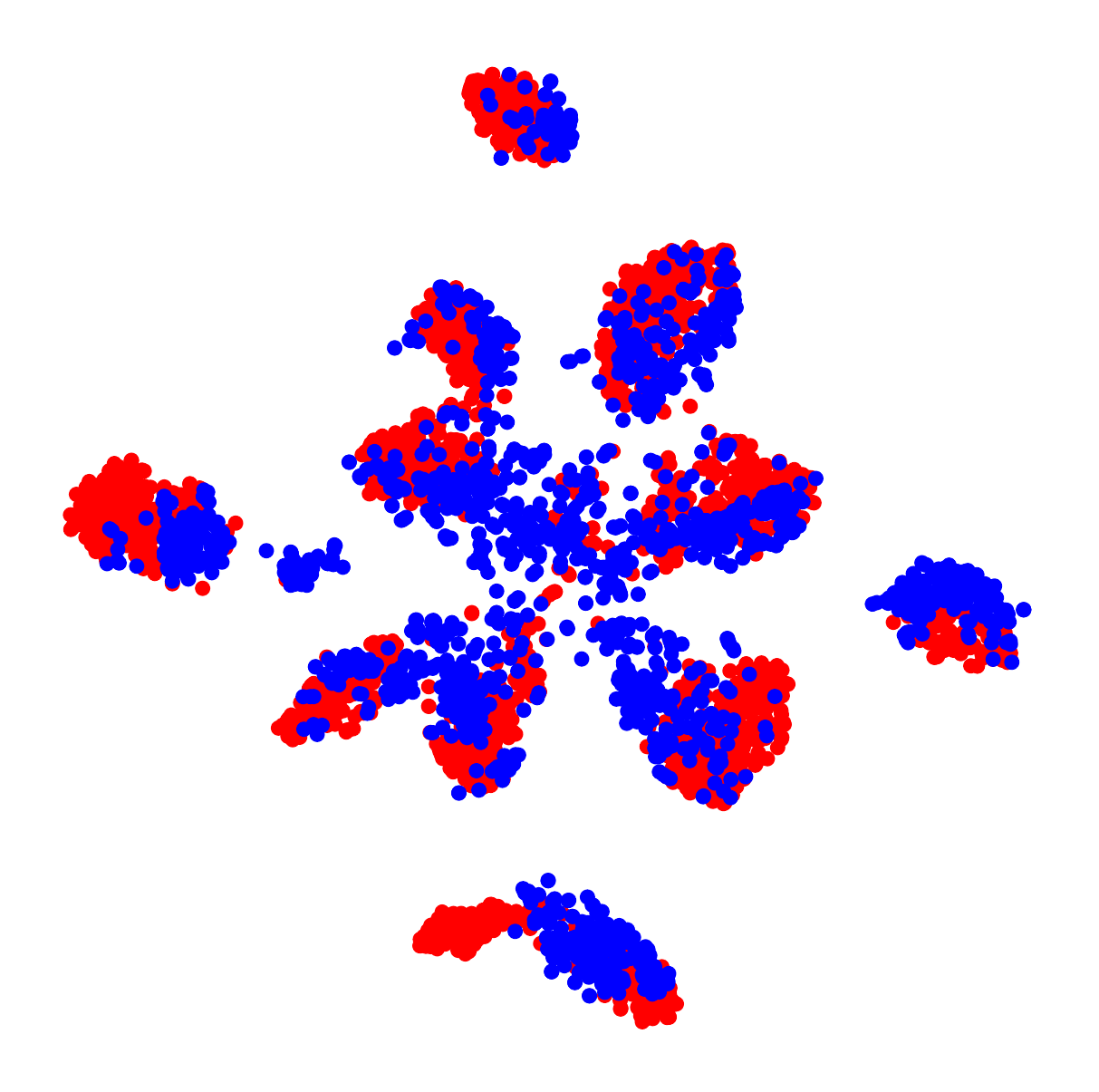}
    }
    \subfigure[]{
    \centering
    \includegraphics[width=1.25in,height=1.0in]{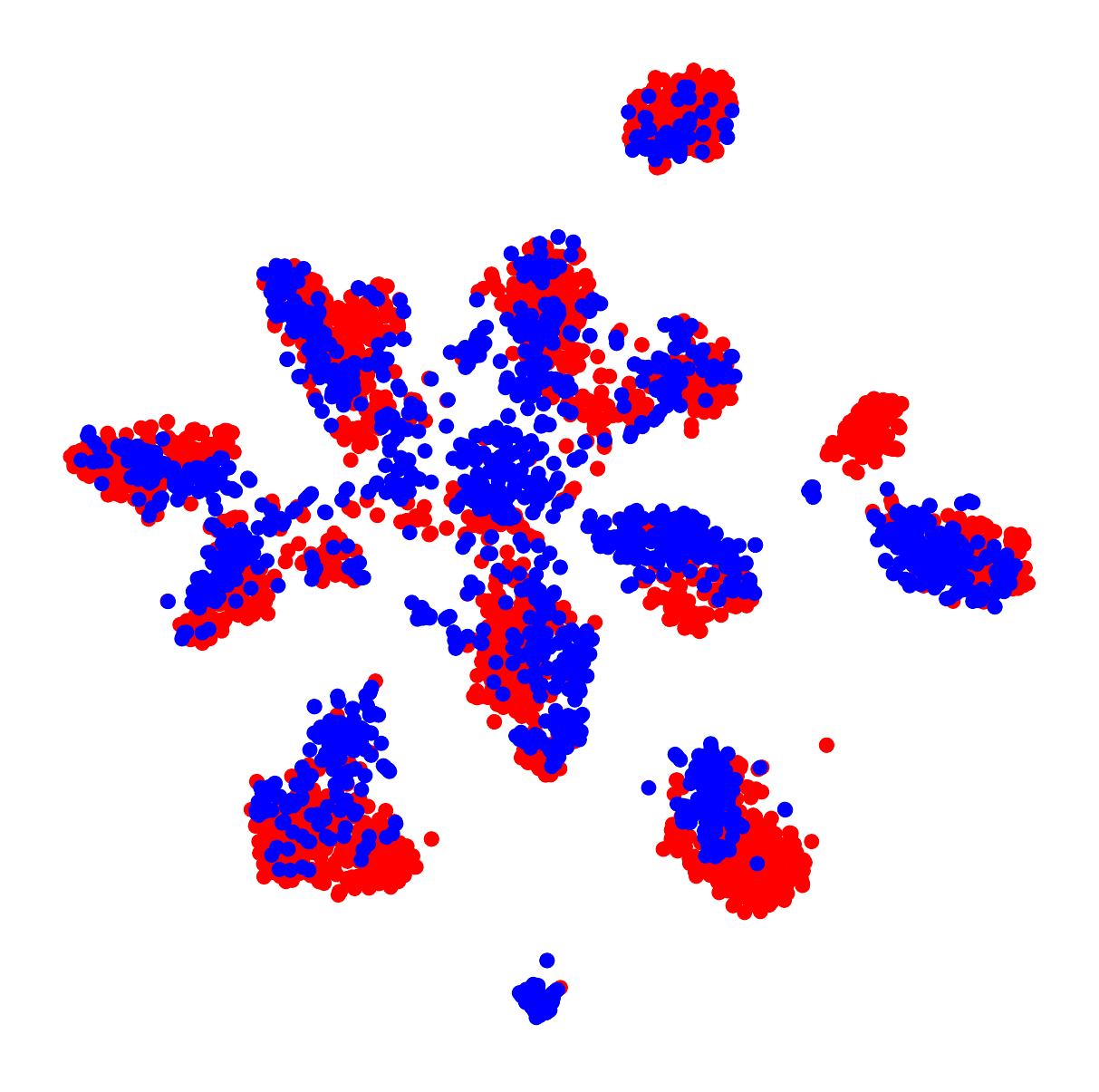}
    }
    \vspace{-3mm}
    
    \subfigure[]{
    \includegraphics[width=1.25in,height=1.0in]{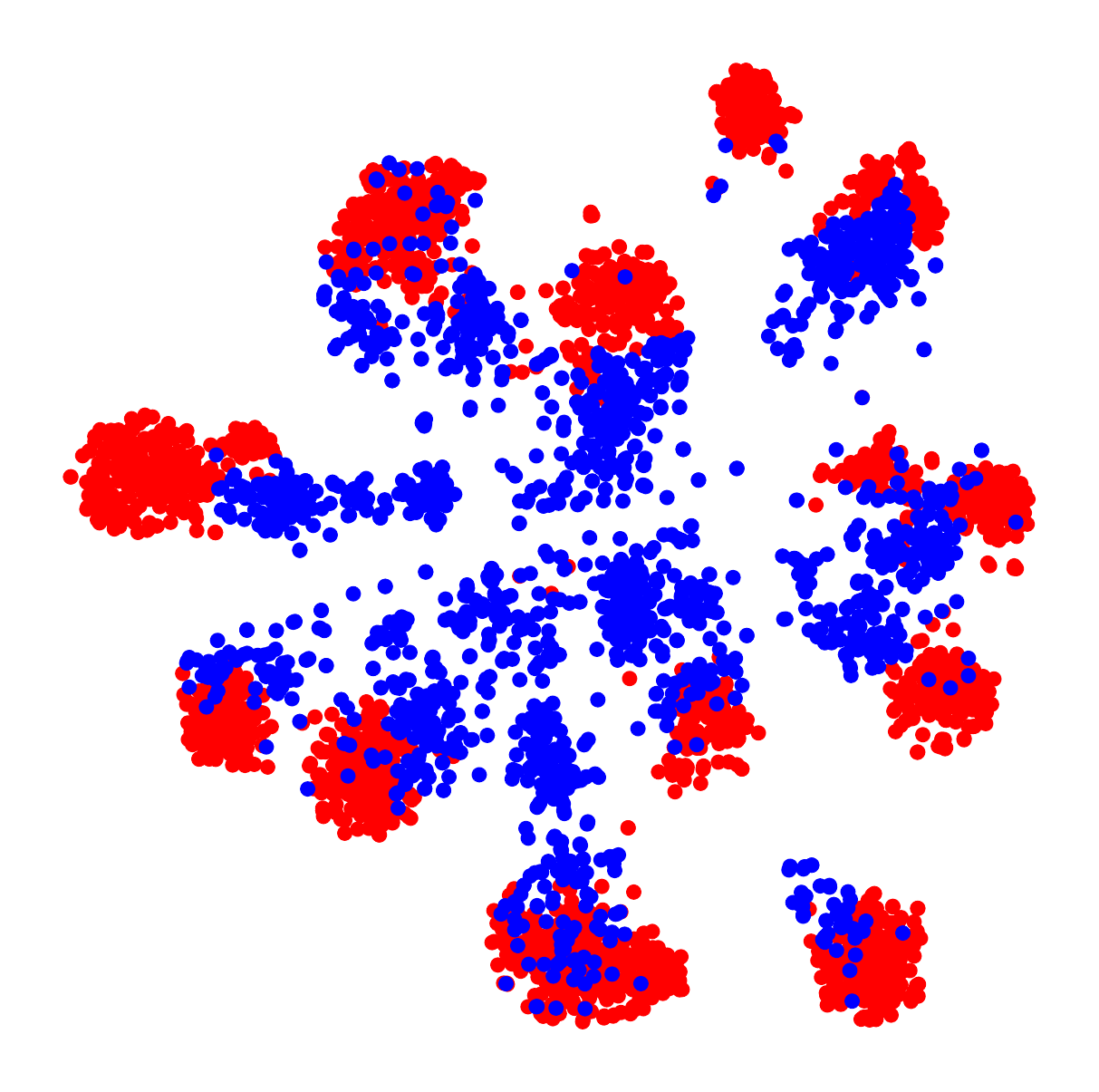}
    }
    \subfigure[]{
    \includegraphics[width=1.25in,height=1.0in]{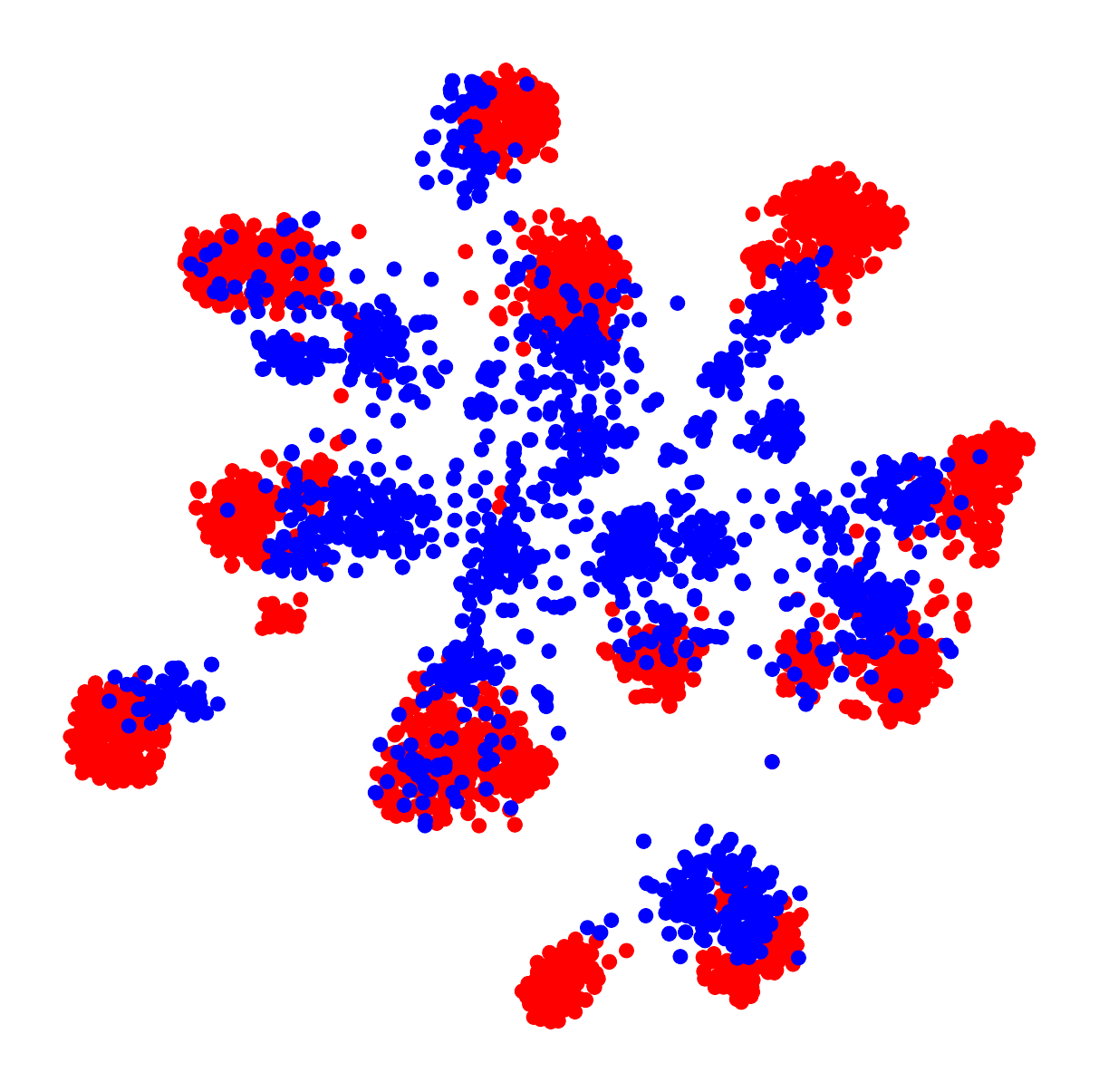}
    }
    \subfigure[]{
    \includegraphics[width=1.25in,height=1.0in]{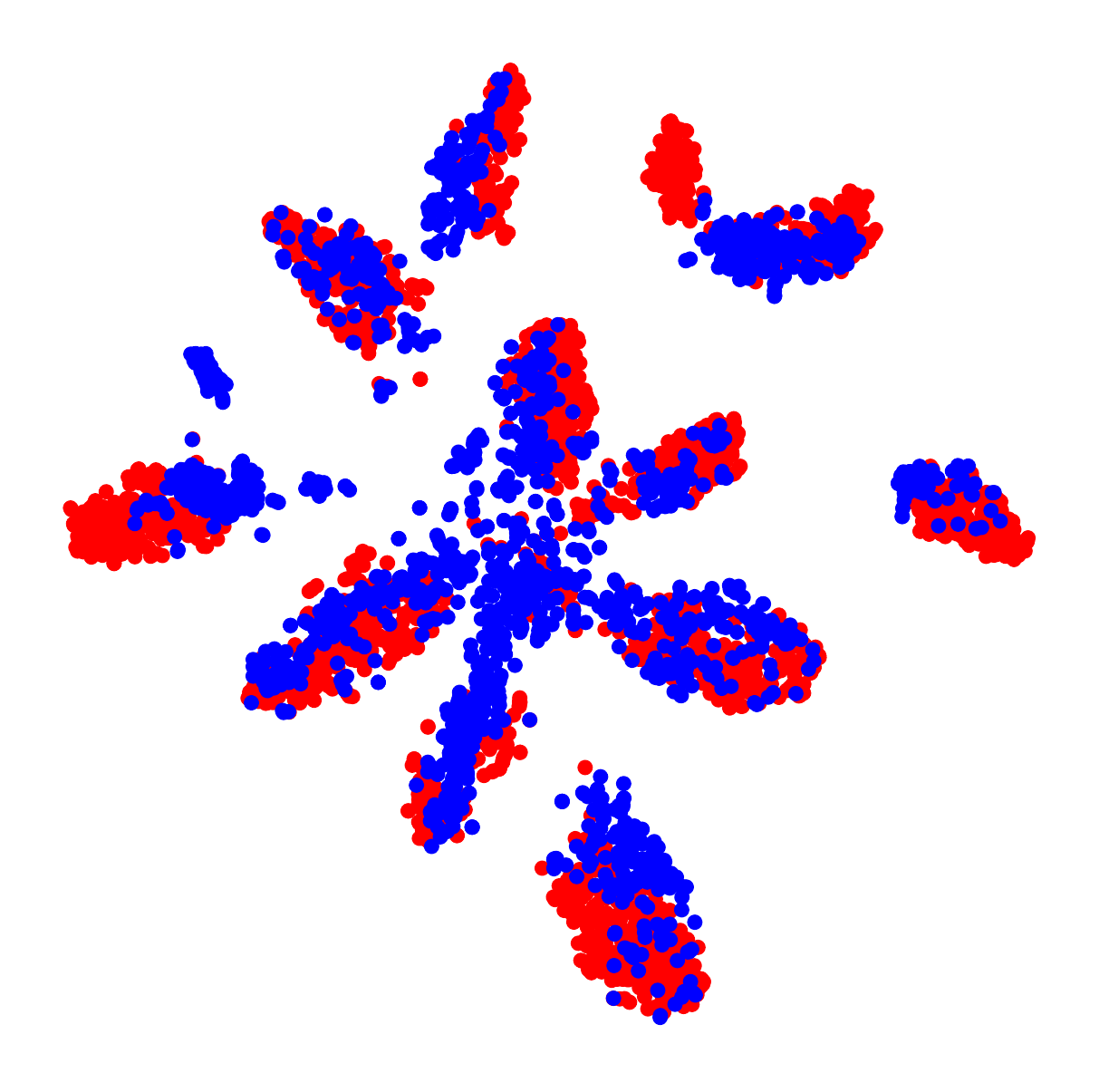}
    }
    \subfigure[]{
    \includegraphics[width=1.25in,height=1.0in]{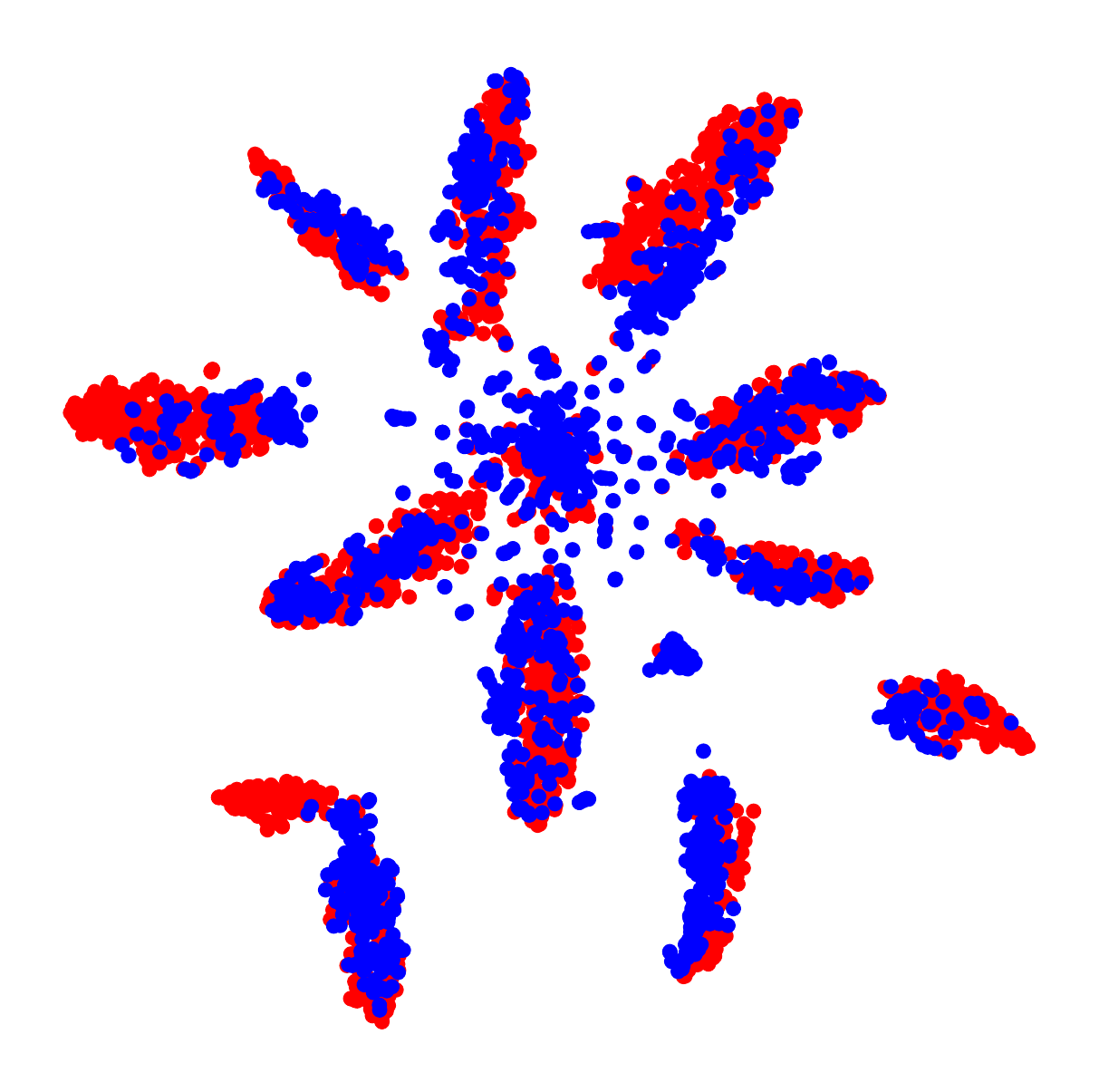}
    }
\vspace{-2mm}
\caption{The t-SNE~\cite{maaten2008visualizing} visualization results of shared ten-class features in the 3-shot \textbf{R}$\rightarrow$\textbf{S} problem  obtained by four \textbf{1-\bm{$C$}}-based methods: (a) S+T, (b) ENT, (c) MME, (d) \bm{$H_{src}$}+\bm{$H_{tar}$}, and four \textbf{2-\bm{$C$}}-based methods: (e) S+T, (f) \bm{$H_{src}$}, (g) \bm{$H_{tar}$}, (h) Ours (\ie, \bm{$H_{src}$}+\bm{$H_{tar}$}). The feature points of source and target domains are indicated by red and blue spots respectively.}\label{f4}
\vspace{-2mm}
\end{figure*}

\textbf{Ablation Study.}
To investigate the effects of each module, we introduce the ablation study on the adaptation scenarios \textbf{R$\rightarrow$S} and  \textbf{R$\rightarrow$C} in Table~\ref{t3} and visualize their features in Fig.~\ref{f4}. As shown in Table~\ref{t3}, there are five different components including one single classifier (i.e., \textbf{1-\bm{$C$}}) which is of the same structure as MME, two classifiers (i.e., \textbf{2-\bm{$C$}}) which consists of the source-scattering and the target-clustering classifier, the target entropy loss (i.e., \bm{${H}_{tar}$}), the combination of the source entropy and the target entropy (i.e., \bm{${H}_{tar}$}+\bm{${H}_{src}$}), and the self-training~\cite{qin2019generatively} (i.e., \textbf{ST}). From the table, we find that the performance has dropped after adding the \bm{${H}_{src}$} to \textbf{1-\bm{$C$}}, which indicates that one classifier is unable to learn opposite structures. However, after introducing \textbf{2-\bm{$C$}} to optimize \bm{${H}_{tar}$}+\bm{${H}_{src}$}, the performance has been significantly improved. Such two observations can strongly prove the necessity of two-classifiers structure for opposite structures learning. The performance of the proposed method can be improved further by introducing pseudo labels. Even without the self-training, our proposed model still outperforms the baseline methods in a large margin.

\begin{table*}[t]
\begin{center}
\caption{ Quantitative results (\%) of ablation study by the backbone ResNet34~\cite{he2016deep}.}\label{t3}
\scalebox{0.8}{
\begin{threeparttable}
 \centering
  \begin{tabular}{|ccccccccc|}
\hline \hline
\multicolumn{5}{|c}{{COMPONENTS}}&\multicolumn{2}{c}{{REAL $\longrightarrow$ SKETCH}}&\multicolumn{2}{c|}{{REAL  $\longrightarrow$  CLIPART}}\\
\multicolumn{1}{|c}{~1-$C$~} &\multicolumn{1}{c}{~2-$C$~}  &\multicolumn{1}{c}{~${H}_{tar}$~} &\multicolumn{1}{c}{${H}_{tar}$+${H}_{src}$}  &\multicolumn{1}{c}{~~ST~~}&ONE-{SHOT} & THREE-{SHOT}
&ONE-{SHOT} & THREE-{SHOT} \\
\hline
{{\Checkmark}} &{ } &{ \Checkmark} &{ } &{ }  &61.03&61.93  &70.04&72.19\\

{\Checkmark} &{ } &{ }  &{\Checkmark } &{ }  &60.40&61.16  &69.24&71.47\\

{ } &{\Checkmark} &{\Checkmark }  &{ } &{ }  &61.11&62.81  &70.53&72.42\\

{ } &{\Checkmark} &{ }  &{ \Checkmark} &{ }  &62.17&63.90  &71.57&74.02\\

{ } &{\Checkmark} &{\Checkmark } &{ } &{\Checkmark}  &61.17&62.58  &70.74&72.83\\

{ } &{\Checkmark} &{ }  &{ \Checkmark} &{\Checkmark}  &\textbf{62.70}&\textbf{64.18}  &\textbf{72.72}&\textbf{75.41}\\
\hline \hline
\end{tabular}
\renewcommand{\labelitemi}{}
\end{threeparttable}
}
\end{center}
\vspace{-4mm}
\end{table*}

\textbf{Convergence Analysis. } We evaluate the convergence of baseline methods as well as our proposed methods on the adaptation scenario \textbf{R$\rightarrow$S} in Fig.~\ref{f5} (a). We can clearly see that the proposed method would reach the highest score around the $5,000$-th epoch and keep stable since that.  

\textbf{Sensitivity of labeled samples. } As shown in Fig.~\ref{f5} (b), we conduct experiments adapted from the \textit{Real} to \textit{Sketch} on the settings of 1-shot, 3-shot and 5-shot SSDA. We notice that our proposed method continuously maintains a leading position compare to other baseline methods under all the settings.

\textbf{Feature Visualization.}
In Fig.~\ref{f4} we employ the t-SNE algorithm~\cite{maaten2008visualizing} to visualize generator features obtained by the methods listed in Table~\ref{t3}. It is obvious that the features learned by our approach are more clustered and separable. In the comparison between (c), (g) and (h), although the \bm{$H_{tar}$}'s features are well clustered, those clusters are not separable enough. \bm{$H_{src}$}+\bm{$H_{tar}$} improves this by expanding decision boundary which implicitly separate multiple target clusters to make cross-domain features more precisely aligned. This case proves the utility of source scattering in our model. 

\textbf{Sensitivity of $\alpha$.} We introduce $\alpha$ to enforce the divergence of two classifiers. As shown in Fig.~\ref{f5} (c)-(d), the best value of $\alpha$ is around 0.75 which means that the slightly different classifiers do help learning opposite structures than the strongly biased (i.e., $\alpha$=0.9 or $\alpha$=0.1) or balanced ones (i.e., $\alpha$=0.5). It proves the necessity of assigning different weights to supervision losses in two classifiers.

\section{Conclusion}

In this paper, we attempt to explore the optimal structure of source domain features and design certain (i.e., scattering) mechanisms to achieve this goal. Specifically, we propose a novel semi-supervised domain adaptation framework inspired by the Unity of Opposites. Our model is composed of a generator network and two classifier networks (i.e., the source-scattering classifier and the target-clustering classifier) designed with the contradictory forms of losses. The source-scattering classifier is applied to disperse the source features and the target-clustering classifier is employed to cluster the target ones. By the cooperation of target-feature clustering and source-feature expansion, the target features would be well enclosed by the expanded source features given a few labeled target samples as support, which results the precise alignment of the cross-domain features. Extensive experiments on DomainNet and Office-home demonstrate the superiority of our approach over the state-of-the-art methods.  Our work could further inspire the community to think what is the better structure of source domain features and how to make it cooperate with target domain structure learning for the unified goal of feature alignment.


\bibliographystyle{splncs04}
\bibliography{egbib}

\clearpage

\title{\Large Supplementary Materials}


\date{}

\maketitle







\section{Theoretical Insights}
In this section, we provide a theoretical analysis for the effectiveness of our proposed model. Based on the theory in \cite{ben2010theory}, we can upper bound the risk on the target domain with the risk on source domain and the domain divergence, \textit{i.e.}, 
\begin{align}
    \forall h\in H, \quad \rr_\ttt(h)\leq \rr_\sss(h)+\frac{1}{2}d_{\hh}(\sss,\ttt)+\delta,
\end{align}
where $\ttt$ and $\sss$ represent the target domain and source domain respectively, $\rr_\ttt(h)$ is the expected risk on domain $\ttt$, $\rr_{\sss}(h)$ is the expected risk on domain $\sss$, $d_{\hh}(p,q)$ represents the $\hh$-distance of distribution $p$ and $q$,  $\delta$ is a constant which is decided by the complexity of the hypothesis space and the error of a perfect hypothesis for both domains. As a consequence, if we train the domain classifiers and the feature extractors with low divergence $d_{\hh}(\sss,\ttt)$, we can get corresponding low risk on the target domain. Now, we will show how our proposed model is connected to this theory. Since $d_{\hh}(\sss,\ttt)$ can be written as: 
\begin{align}\label{divergence}
    d_\hh(\sss,\ttt)=2\sup_{h\in \hh} \left| \Pr_{\bb f^s \sim p }[h(\bb f^s)=1]- \Pr_{\bb f^t \sim q }[h(\bb f^t)=1]\right|,
\end{align}
where $\bb f^s$ and $\bb f^t$ represents the features extracted from domain $\sss$ and domain $\ttt$ respectively. In our model, we use the entropy function $H(\cdot)$ with respect to target domain and source domain to train the parameters of $\ms F_1(\cdot)$, $\ms F_2(\cdot)$ and $\ms G(\cdot)$. Though the entropy function is not the usual classification loss, our model can also be considered as minimizing divergence (\ref{divergence}) via adversarial training strategy on domain $\ttt$ and domain $\sss$ respectively. Let $h$ be a binary classifier whose label is decided by the value of the corresponding entropy function:
\begin{align}
    h(f)=\begin{cases}
    1 & {\rm if}\quad  H(\ms F_i(\bb f)) \geq \gamma\\
    0 & {\rm otherwise}
    \end{cases},
\end{align}
where $i=1,2$, and $\gamma$ is a threshold of the classifier. To facilitate analysis, we just assume the output of the classifiers $\ms F_1(\cdot)$ and $\ms F_2(\cdot)$ are the conditional probabilities. In this case, we can obtain the upper bound of $d_\hh(\sss,\ttt)$ as following:
\begin{align}\label{divergence2}
    d_\hh(\sss,\ttt) &\approx 2\sup_{\ms F_1, \ms F_2} \left| \Pr_{\bb f^s \sim p }[H(\ms F_1(\bb f^s))\geq \gamma] \right.\nonumber\\
    &\left.\quad - \Pr_{\bb f^u \sim q }[H(\ms F_2(\bb f^u))\geq \gamma]\right| \nonumber \\
    & = 2\sup_{\ms F_1, \ms F_2} \left( \Pr_{\bb f^u \sim q }[H(\ms F_2(\bb f^u))\geq \gamma] \right. \nonumber\\
    &\quad \left.-\Pr_{\bb f^s \sim p }[H(\ms F_1(\bb f^s))\geq \gamma]\right), 
\end{align}
where the approximate equality is due to the number of the labeled samples on target domain is much smaller than unlabeled ones, hence we can use the probability on unlabeled samples to replace the probability on whole target domain. The equality above is due to  the assumption that $\Pr_{\bb f^u \sim q }[H(\ms F_2(\bb f^u))\geq \gamma]\geq \Pr_{\bb f^s \sim p }[H(\ms F_1(\bb f^s))\geq \gamma]$, which is reasonable since we have access to the labels of all the data in the source domain, which means that we can make the corresponding entropy be $0$. Replace $\sup$ with $\max$ in (\ref{divergence2}), we can rewrite it as:

\begin{align*}
     d_\hh(\sss,\ttt)&\approx  2\max_{\ms F_1, \ms F_2} \left( \Pr_{\bb f^u \sim q }[H(\ms F_2(\bb f^u))\geq \gamma] \right.\\
     &\quad- \left. \Pr_{\bb f^s \sim p }[H(\ms F_1(\bb f^s)) \geq \gamma]\right ) \nonumber \\ 
     & = 2\min_{\ms F_1, \ms F_2} \left(-\Pr_{\bb f^u \sim q }[H(\ms F_2(\bb f^u))\geq \gamma] \right.\\
     &\left. \quad+\Pr_{\bb f^s \sim p }[H(\ms F_1(\bb f^s))\geq \gamma]\right ) \nonumber \\
     & = -2 \min_{\ms F_1, F_2} \Pr_{\bb f^u \sim q }[H(\ms F_2(\bb f^u))\geq \gamma]\\
     &\quad+ 2\min _{\ms F_1, F_2} \Pr_{\bb f^s \sim p }[H(\ms F_1(\bb f^s))\geq \gamma] \nonumber\\
    & = -2 \min_{\ms F_2} \Pr_{\bb f^u \sim q }[H(\ms F_2(\bb f^u))\geq \gamma] \\
    &\quad+ 2\min _{\ms F_1} \Pr_{\bb f^s \sim p }[H(\ms F_1(\bb f^s))\geq \gamma],
\end{align*}

which exactly matches with the update rules (\ref{F_1}), (\ref{F_2}) in our model. Intuitively, we can get this upper bound of $d_\hh(\sss,\ttt)$ via finding $\ms F_2$ which produces the maximum of the entropy loss on target domain and finding $\ms F_1$ which produces the minimum of the entropy loss on source domain. Moreover, we aim to minimize the divergence with respect to the features $\bb f^s$ and $\bb f^u$ to bound the risk on $\ttt$: \begin{align}\label{problem}
 &\min_{\bb f^s,\bb f^u}\left\{ \max_{\ms F_1, \ms F_2} \left( 2 \Pr_{\bb f^u \sim q }[H(\ms F_2(\bb f^u))\geq \gamma] \nonumber \right.\right.\\
 & \left.\left.-2\Pr_{\bb f^s \sim p }[H(\ms F_1(\bb f^s))\geq \gamma]\right )\right\},
\end{align}
where finding minimum with respect to $\bb f^s$ and $\bb f^u$ is equivalent to finding the feature extractor $\ms G(\cdot)$ to achieve that minimum, which corresponds to the step (\red{9}) in our model. Therefore, via iteratively train $\ms F_1(\cdot)$, $\ms F_2(\cdot)$ and $\ms G(\cdot)$, we approximate the optimal solution for problem (\ref{problem}). In other words, we can minimize the divergence $d_\hh(\sss,\ttt)$ to effectively reduce the risk on the target domain $\ttt$.

\section{Benchmark Datasets}

\begin{figure*}[t]
\centering
\scalebox{1}{\includegraphics[width=0.975\textwidth, height=55mm]{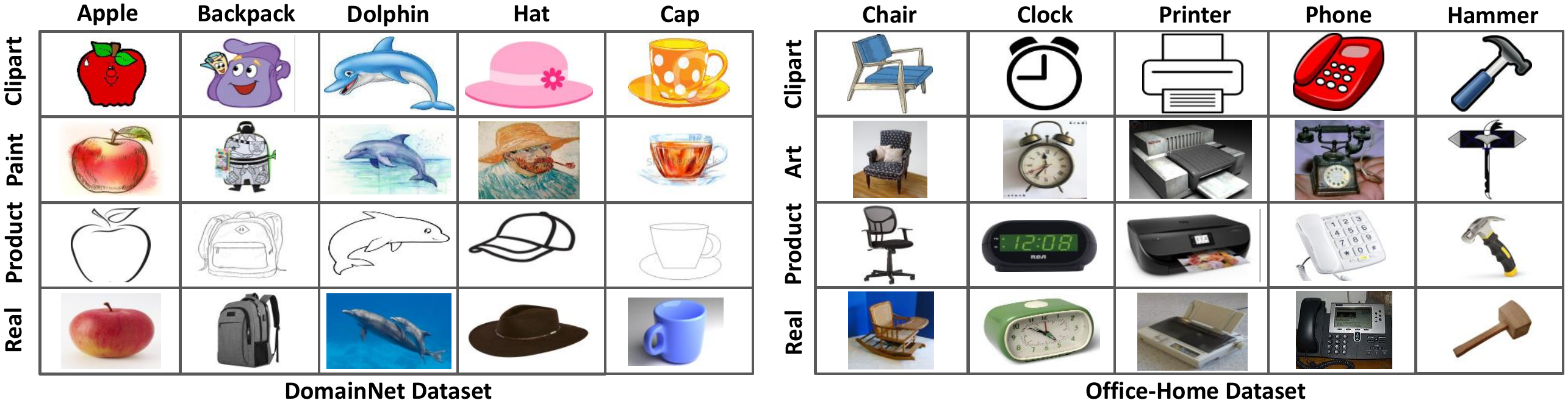}}
\caption{Example images of DomainNet~\cite{peng2019moment} and Office-home~\cite{venkateswara2017deep} Datasets. Compared with the Office-home, DomainNet is a recent dataset involved with more examples} \label{sup:f1}
\end{figure*}

Our proposed approach is evaluated on the latest DA benchmarks, e.g., DomainNet\footnote{\url{http://ai.bu.edu/M3SDA/}}~\cite{peng2019moment} and Office-home\footnote{\url{http://hemanthdv.org/OfficeHome-Dataset/}}~\cite{venkateswara2017deep}, which are shown in Fig.~\ref{sup:f1}. DomainNet is a multi-source domain adaptation benchmark containing $6$ domains and about $600,000$ images among $345$ categories. For a fair evaluation, we take the same protocol of MME~\cite{saito2019semi} where $4$ domains including \textit{Real} (\textbf{R}), \textit{Clipart} (\textbf{C}), \textit{Painting} (\textbf{P}) and \textit{Sketch} (\textbf{S}) with $126$ classes picked for evaluation. The Office-home dataset is a well explored UDA benchmark which consists of 4 domains including \textit{Real} (\textbf{R}), \textit{Clipart} (\textbf{C}), \textit{Art} (\textbf{A}) and \textit{Product} (\textbf{P}) with $65$ classes.

\section{Visualization}
\begin{figure*}[t]
    \centering
    \subfigure[]{
    \centering
    \includegraphics[width=1.2in,height=0.9in]{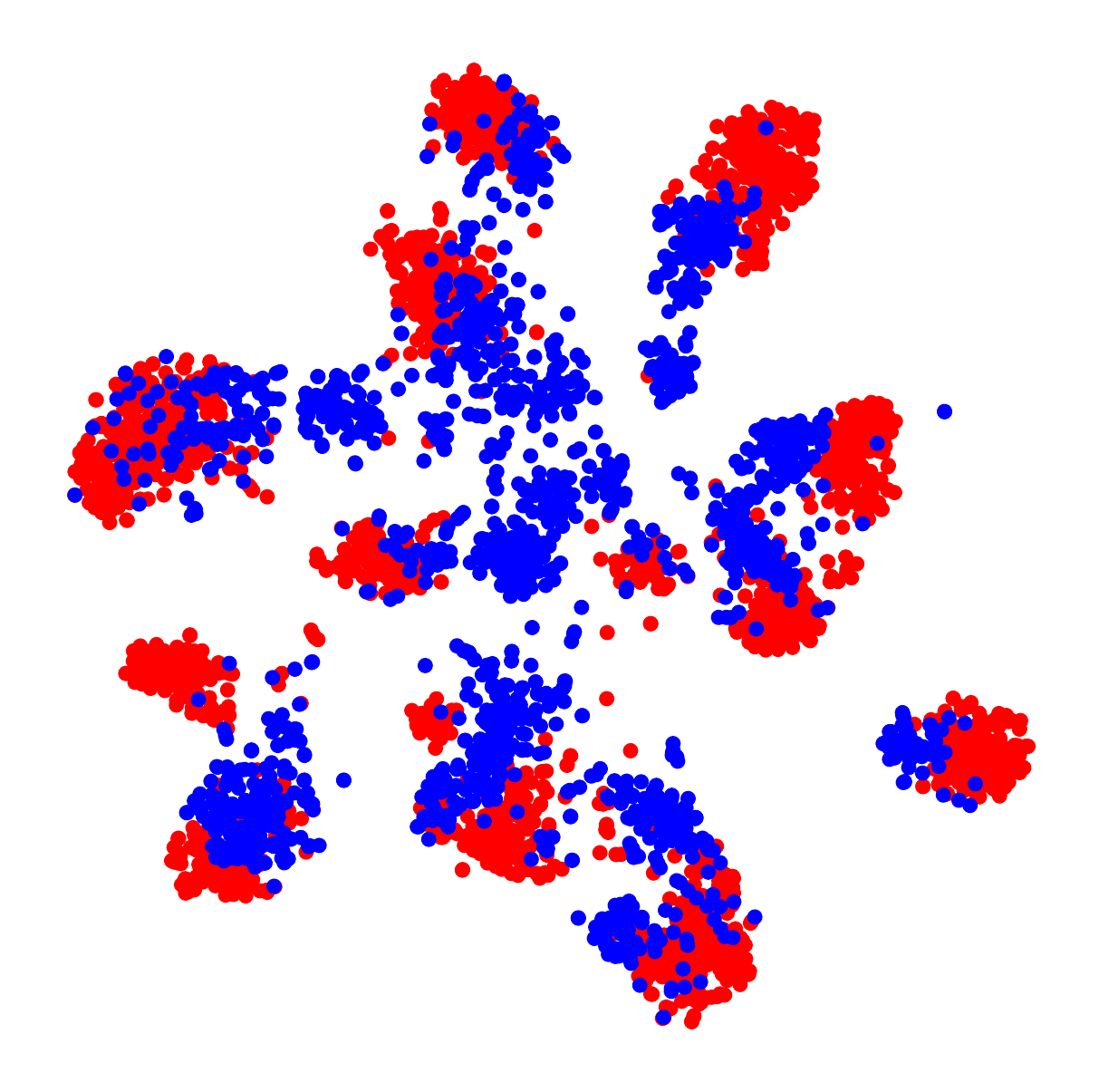}
    }
    \subfigure[]{
    \centering
    \includegraphics[width=1.2in,height=0.9in]{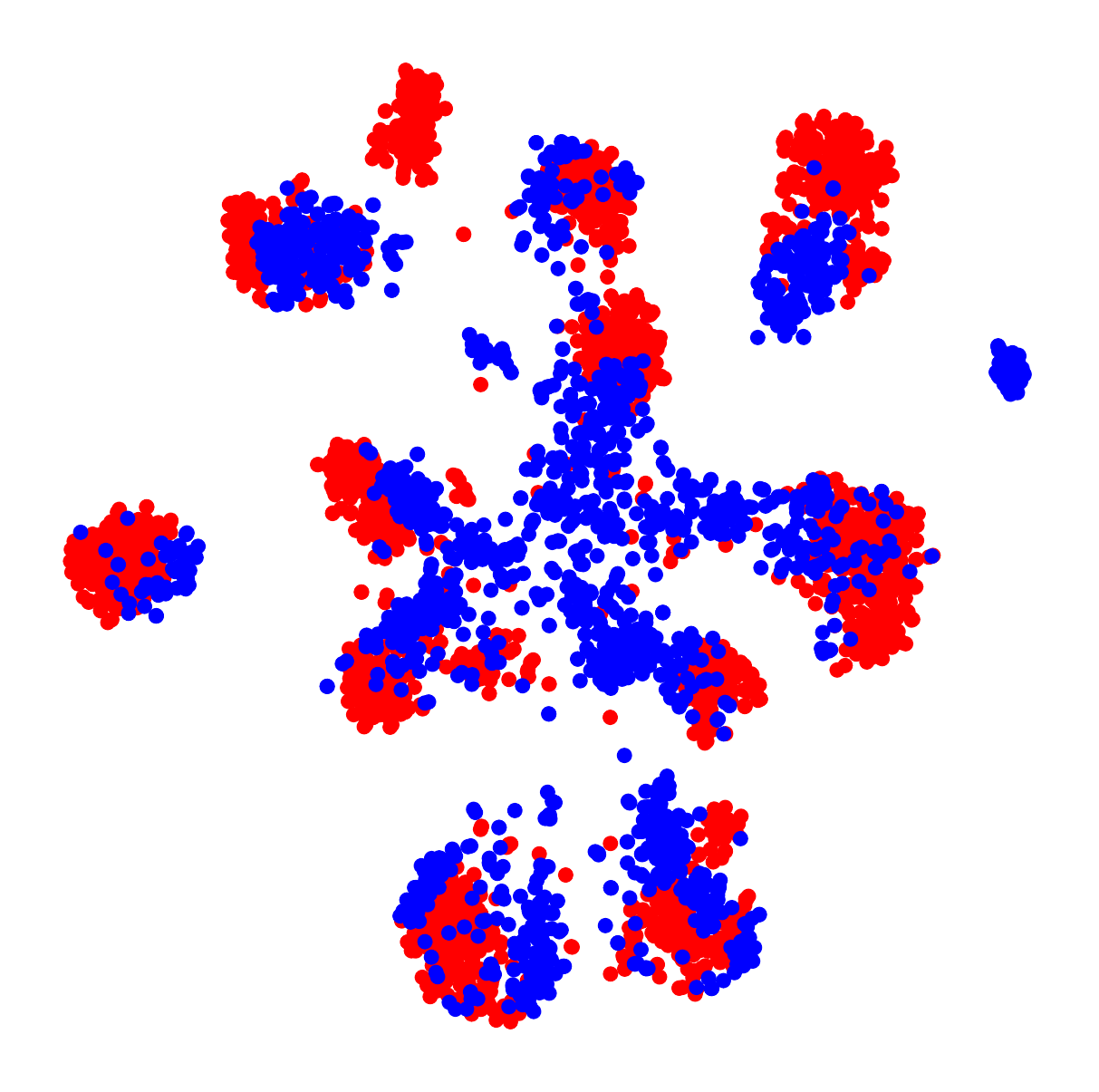}
    }
    \subfigure[]{
    \centering
    \includegraphics[width=1.2in,height=0.9in]{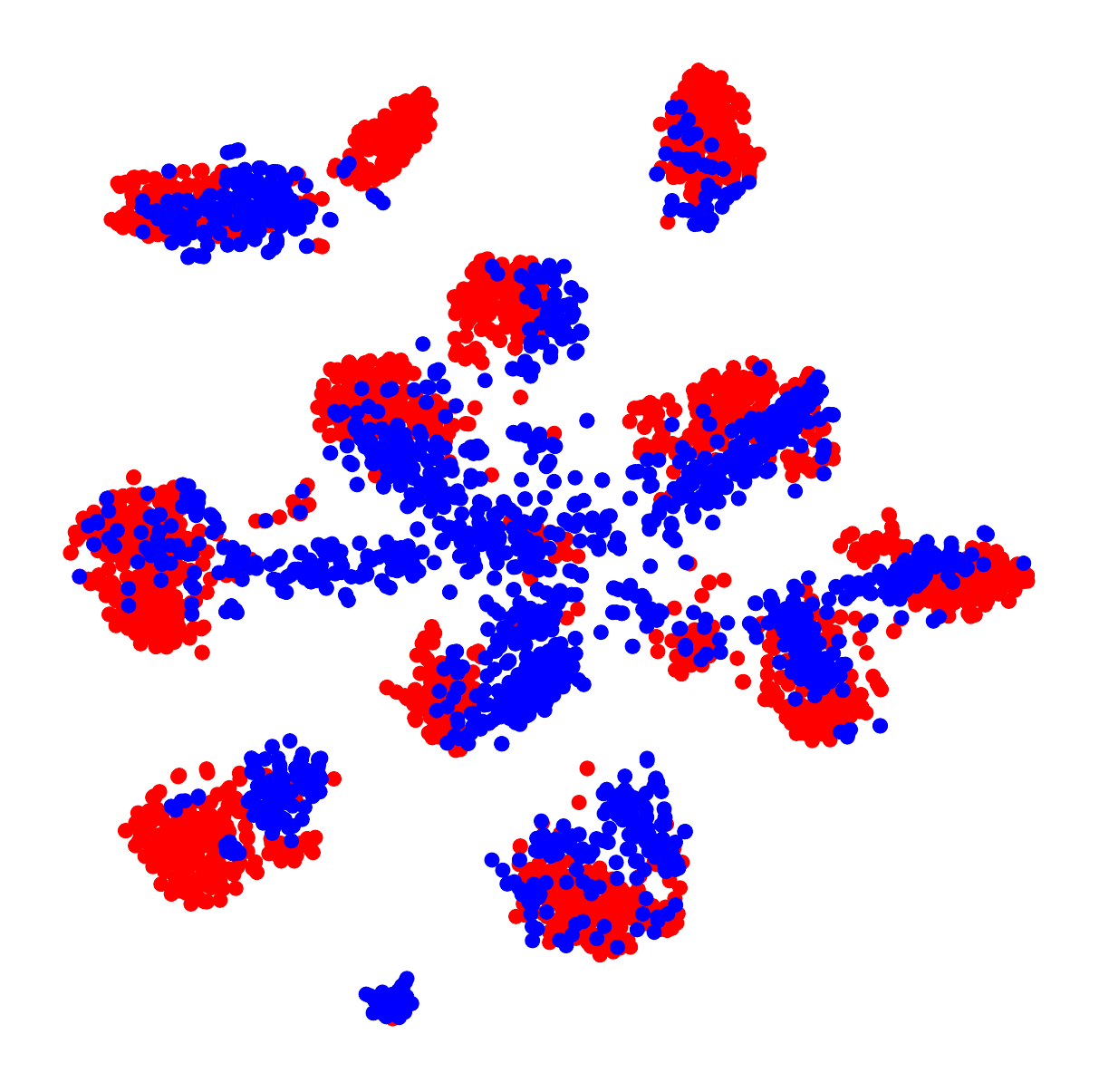}
    }
    \subfigure[]{
    \centering
    \includegraphics[width=1.2in,height=0.9in]{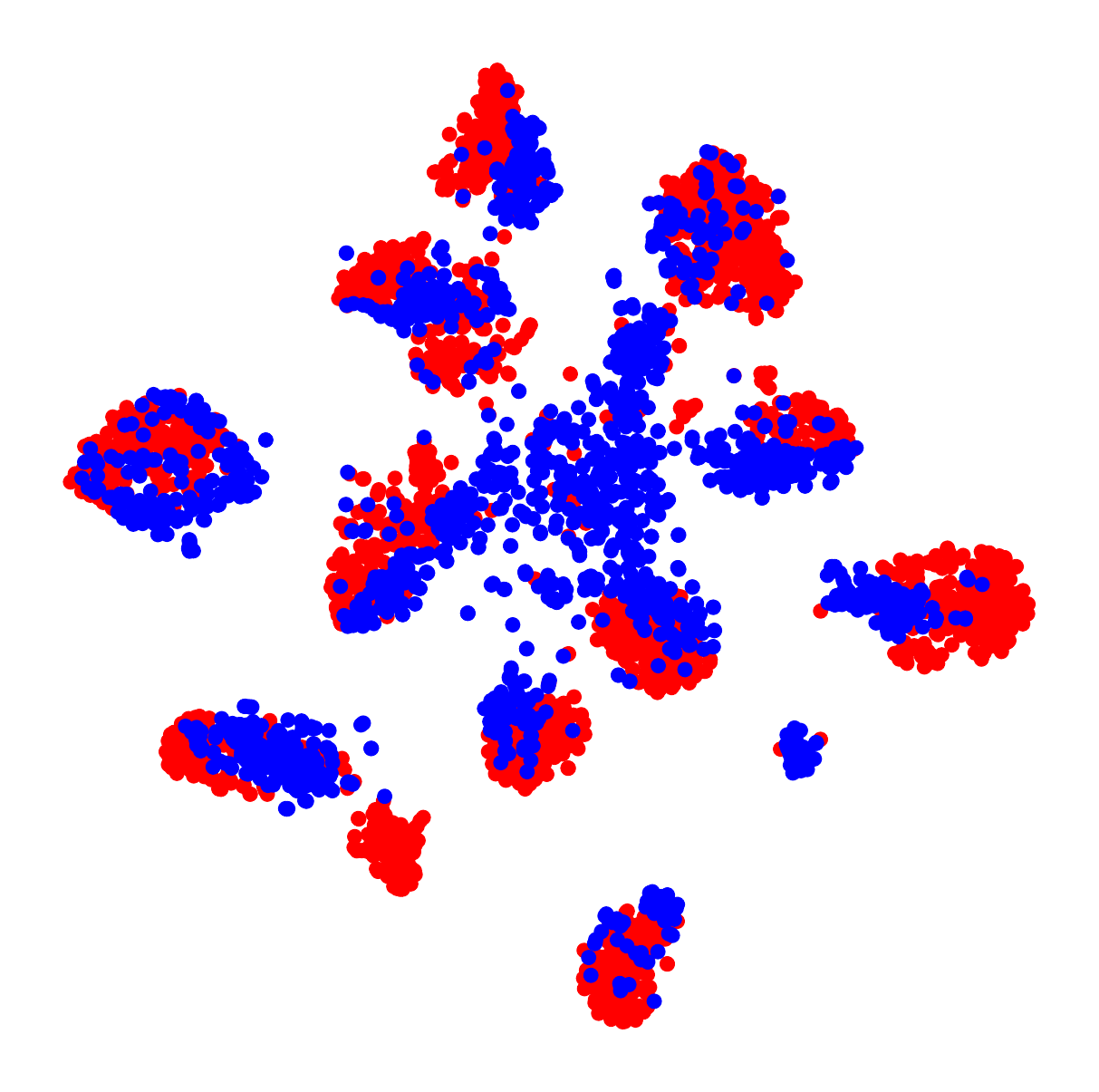}
    }
    \subfigure[]{
    \centering
    \includegraphics[width=1.2in,height=0.9in]{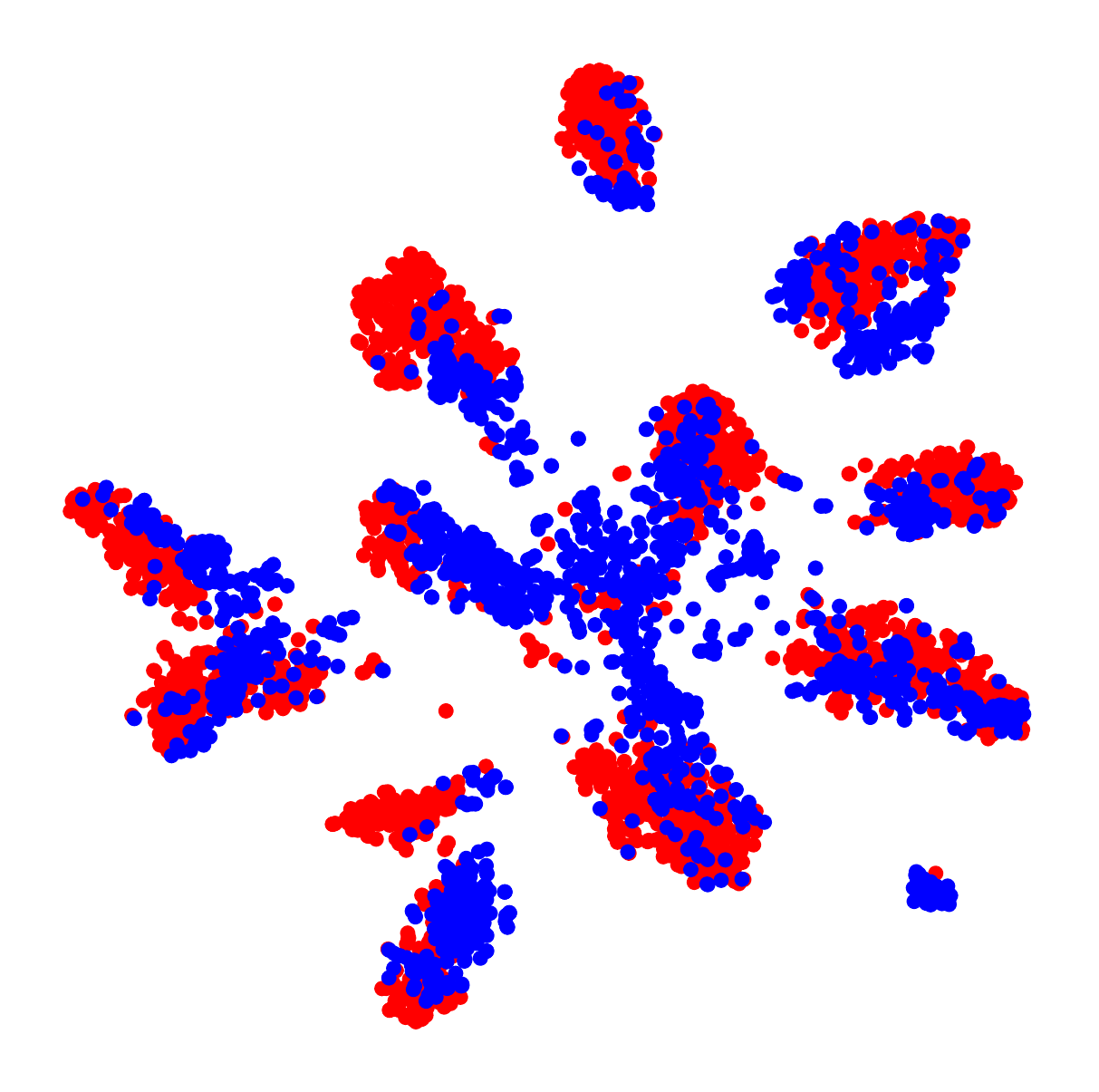}
    }
    
    \subfigure[]{
     \includegraphics[width=1.2in,height=0.9in]{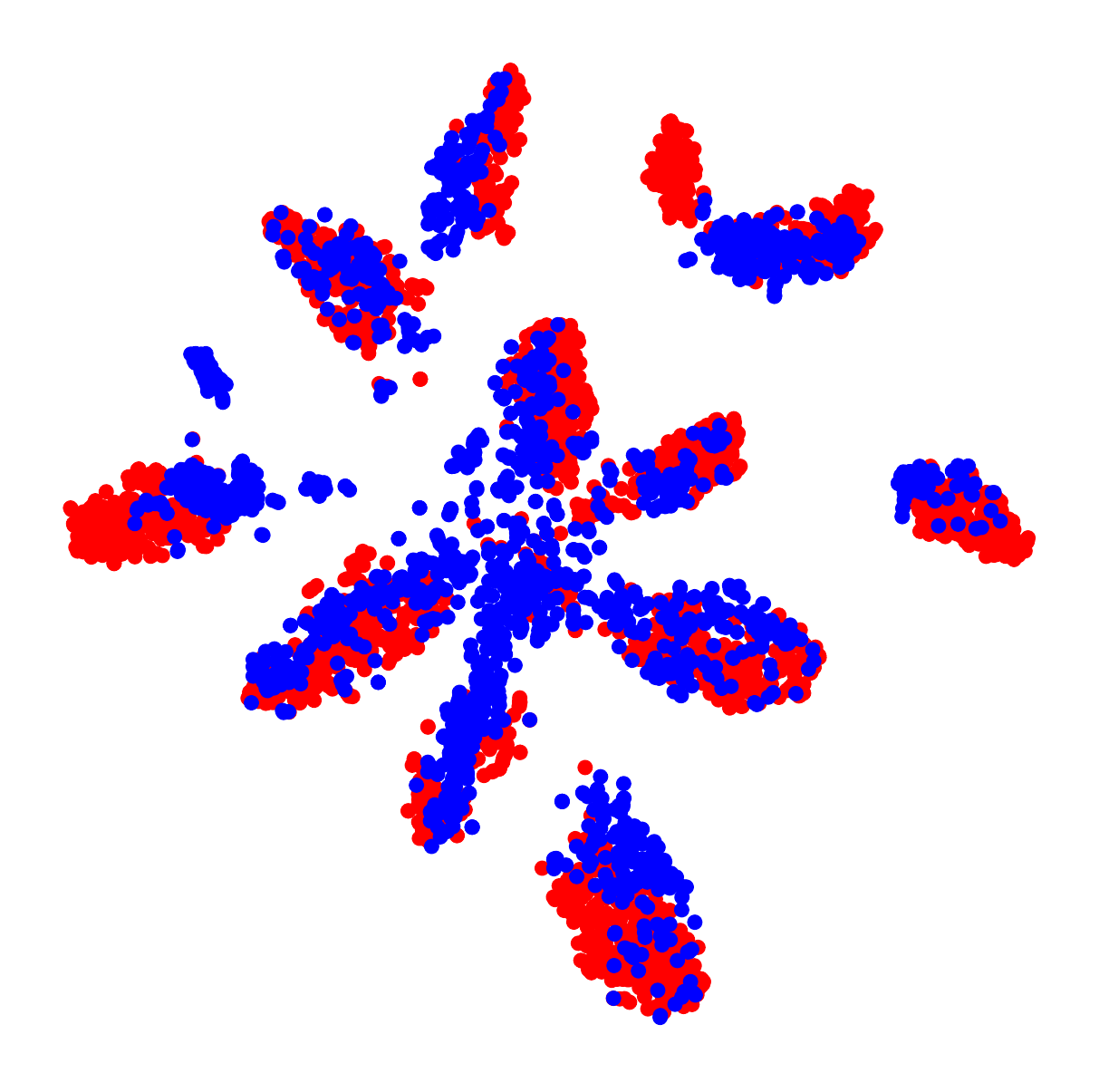}
    }
    \subfigure[]{
     \includegraphics[width=1.2in,height=0.9in]{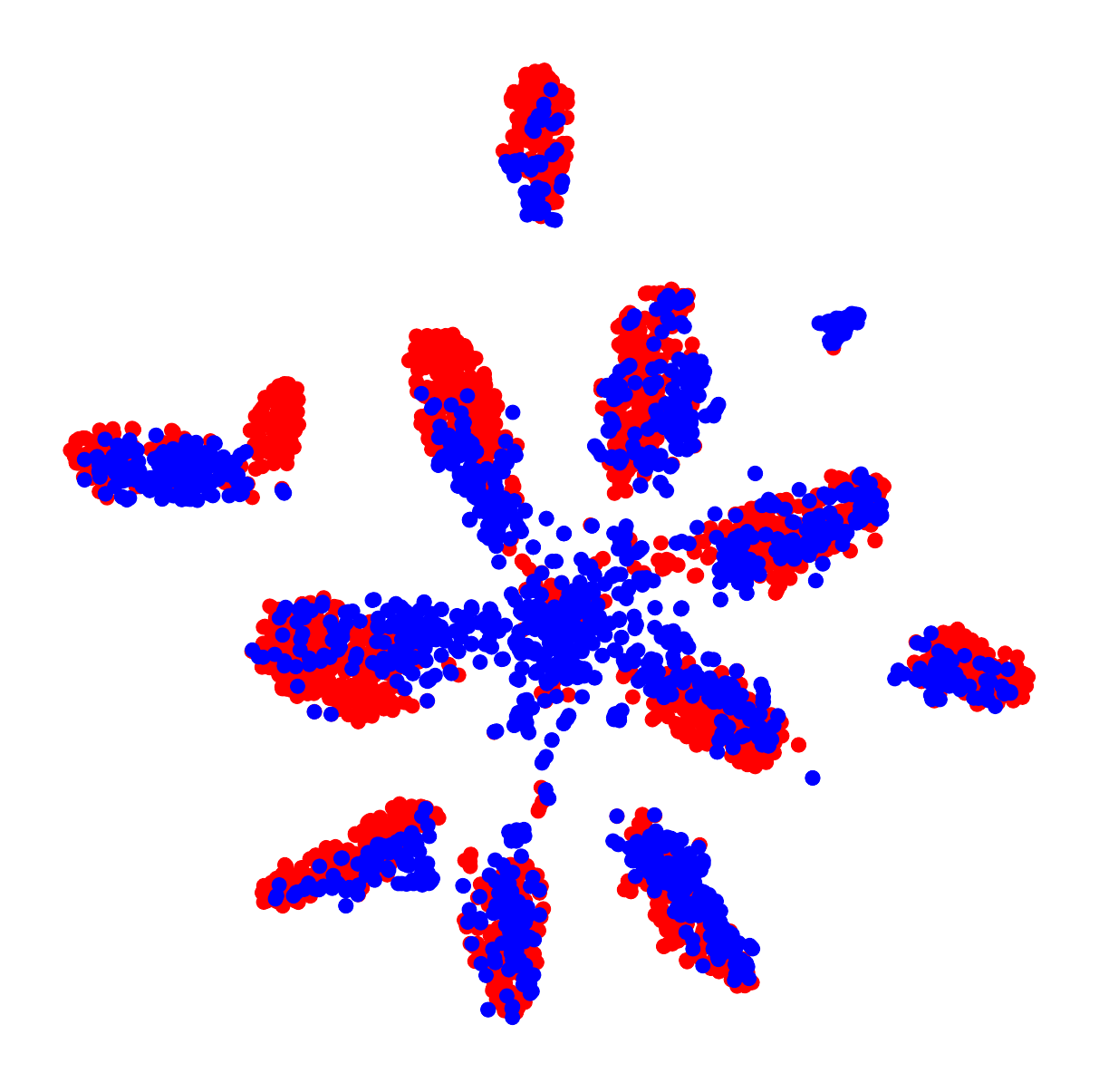}
    }
    \subfigure[]{
     \includegraphics[width=1.2in,height=0.9in]{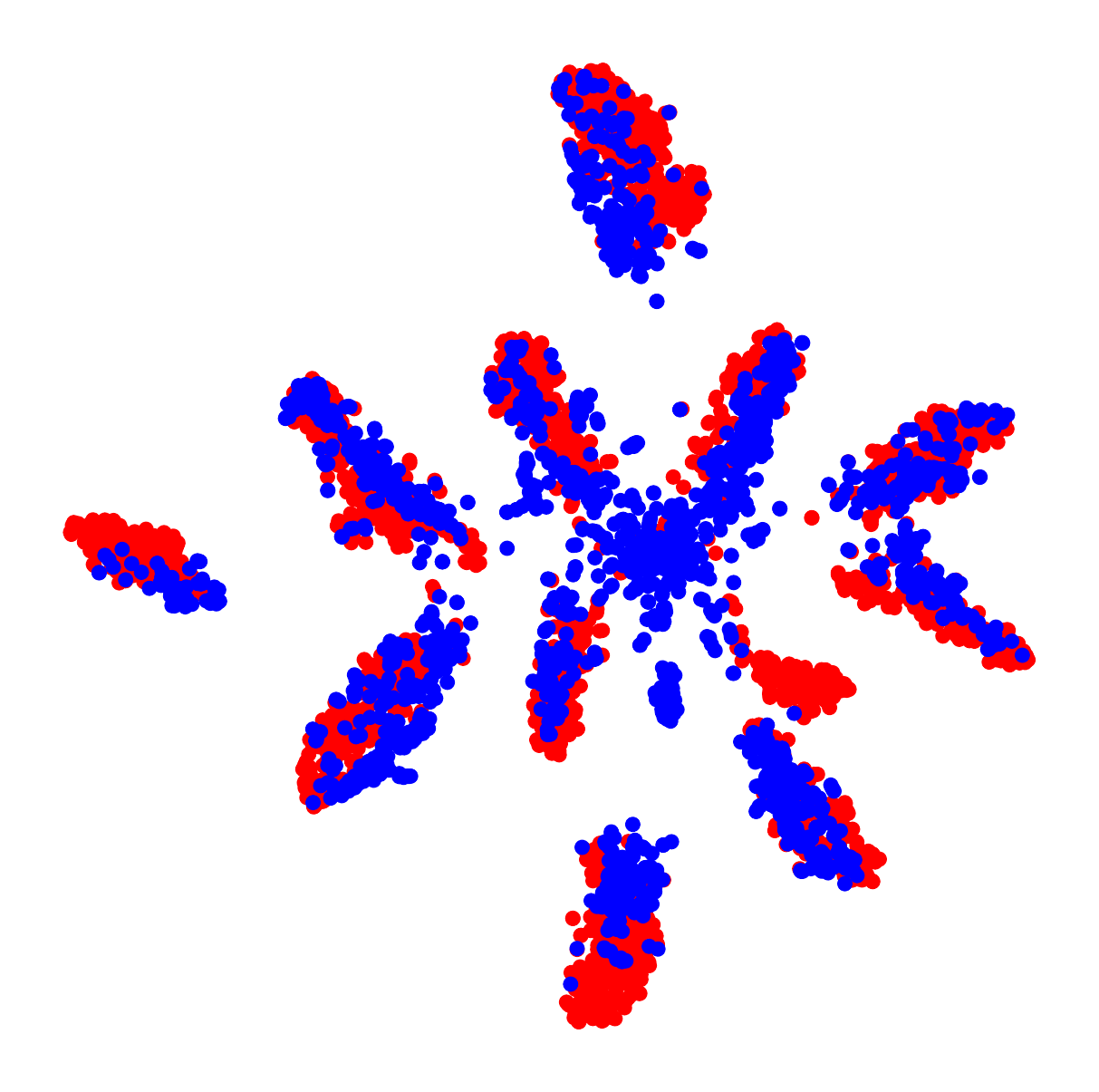}
    }
    \subfigure[]{
     \includegraphics[width=1.2in,height=0.9in]{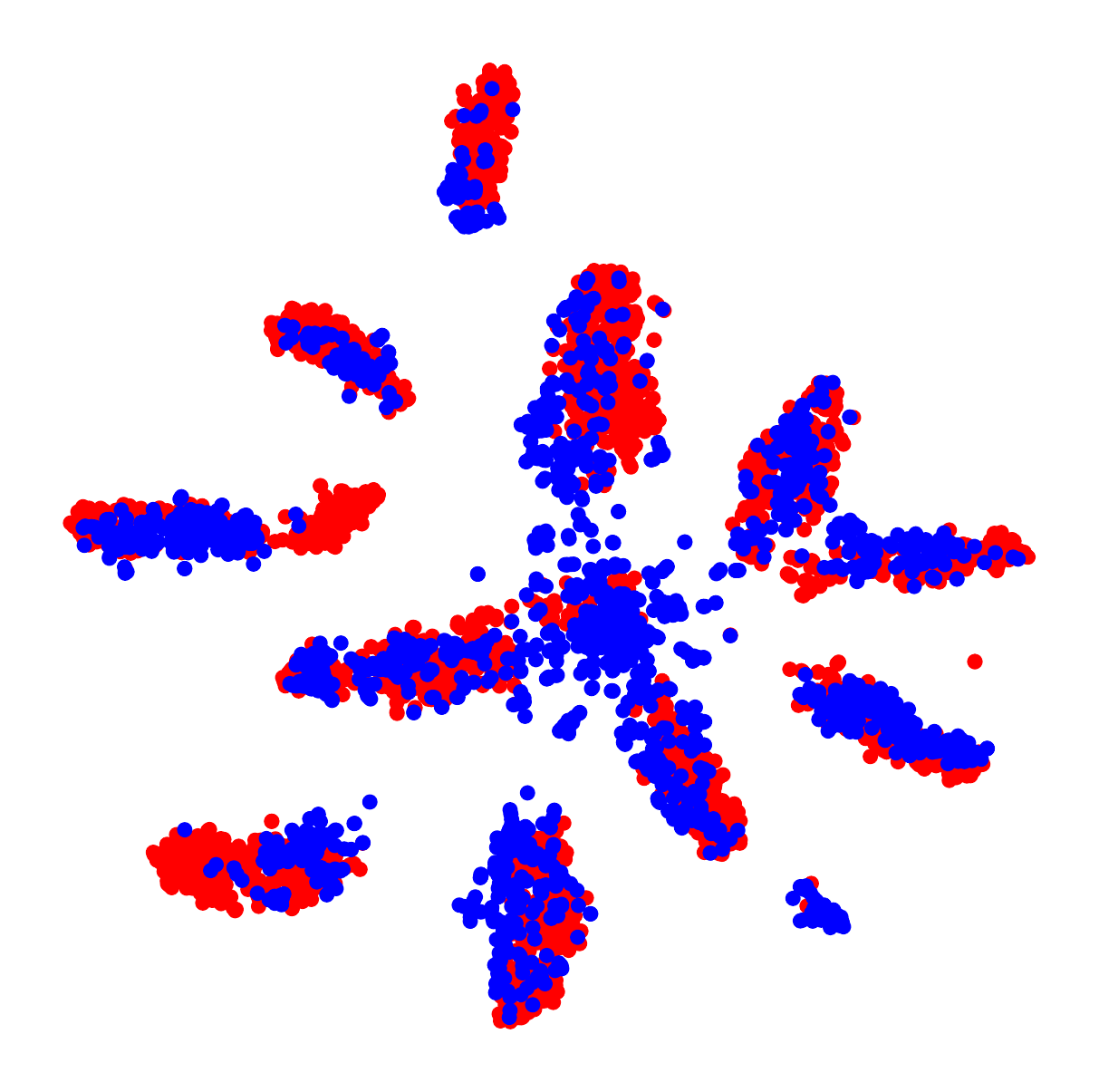}
    }
    \subfigure[]{
     \includegraphics[width=1.2in,height=0.9in]{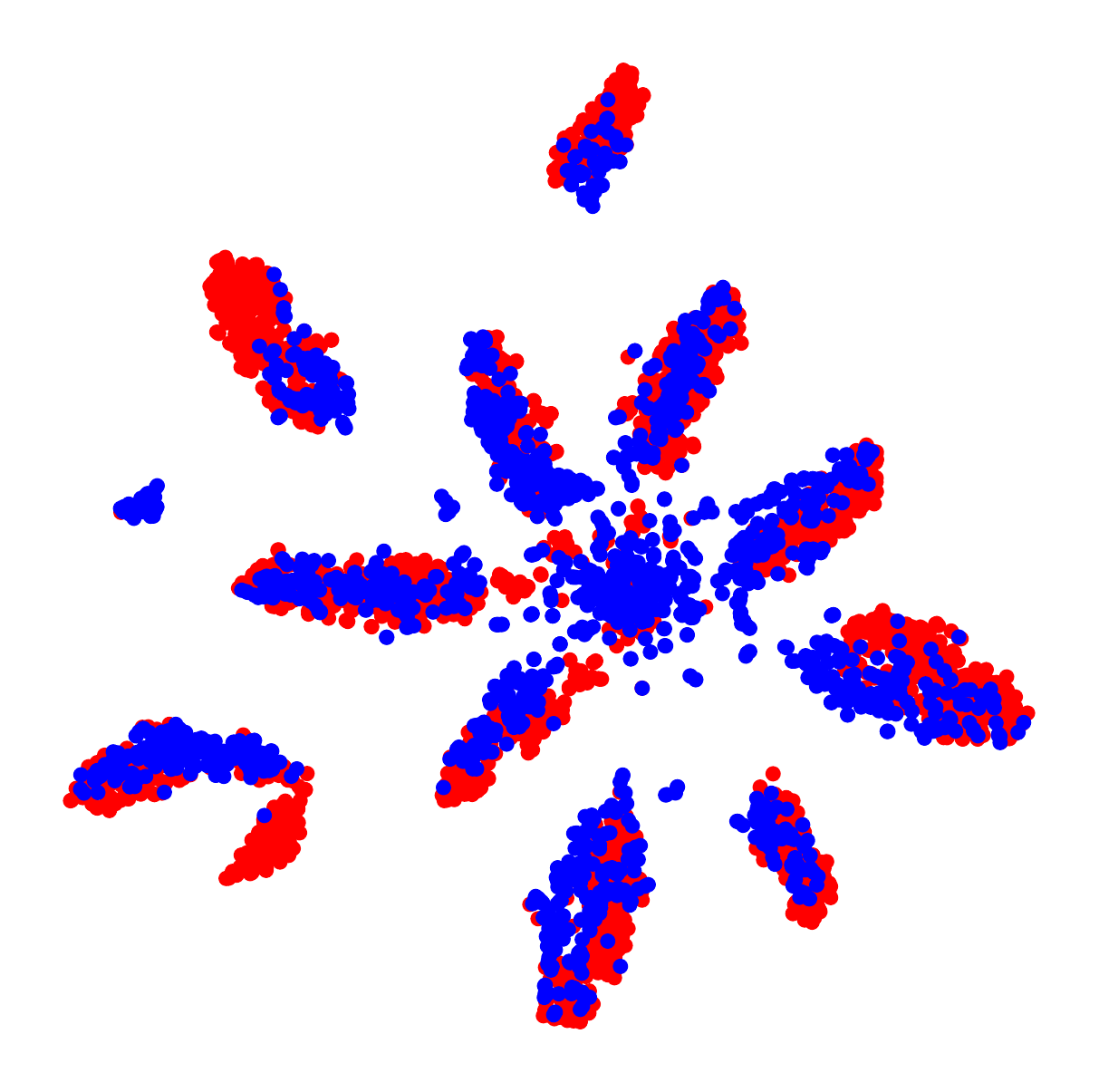}
    }
\caption{The t-SNE~\cite{maaten2008visualizing} visualization results of shared ten-class features in the 3-shot \textbf{R$\rightarrow$S} problem  obtained by \bm{$H_{tar}$} methods at the: (a) 1000-th, (b) 2000-th, (c) 3000-th, (d) 4000-th, (e) 5000-th, (f) 6000-th, (g) 7000-th, (h) 8000-th, (i) 9000-th, and (j) 10000-th epoch.}\label{sup:f5}
\end{figure*}

\begin{figure*}[t]
    \centering
    \subfigure[]{
    \centering
    \includegraphics[width=1.2in,height=0.9in]{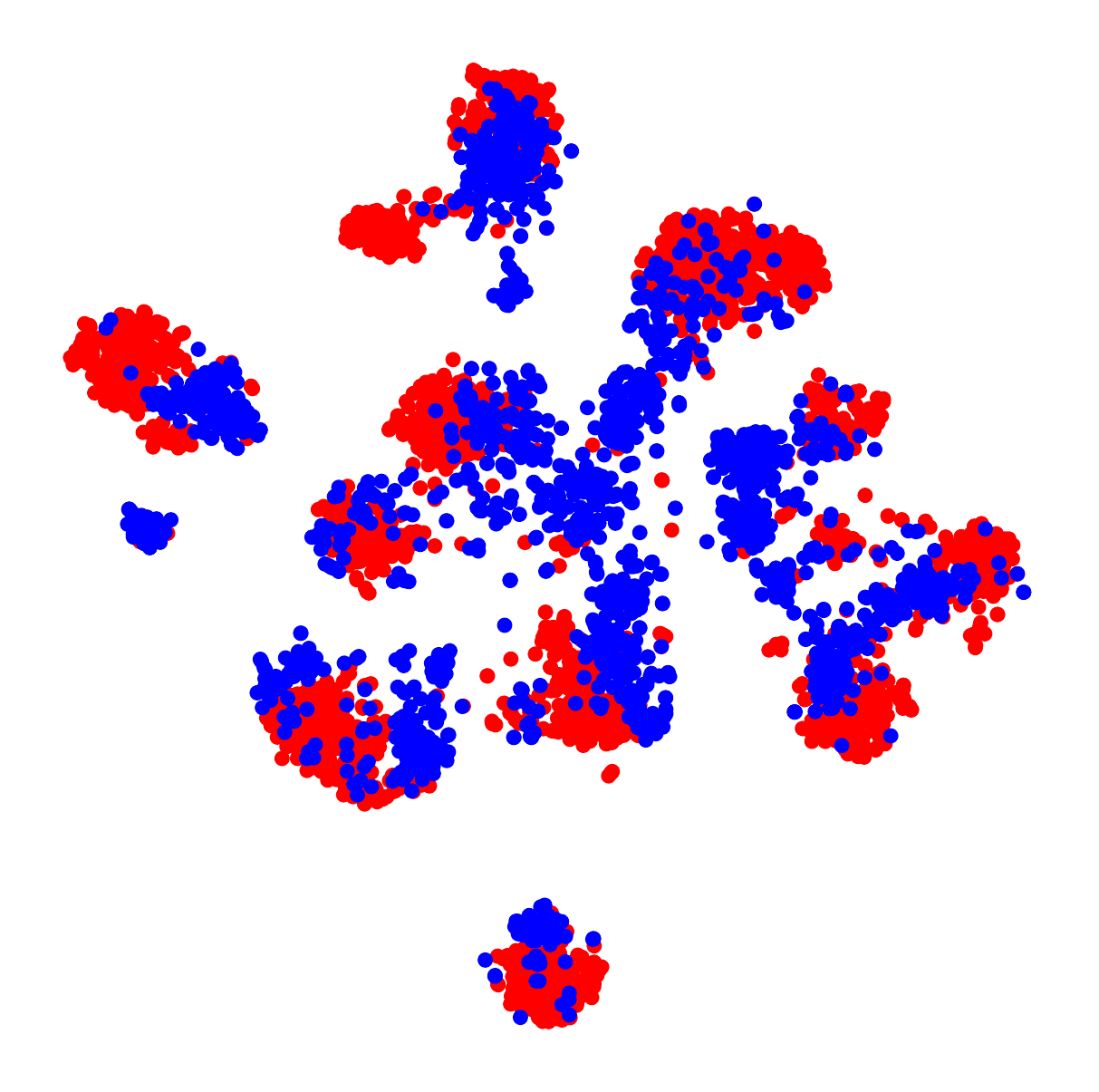}
    }
    \subfigure[]{
    \centering
    \includegraphics[width=1.2in,height=0.9in]{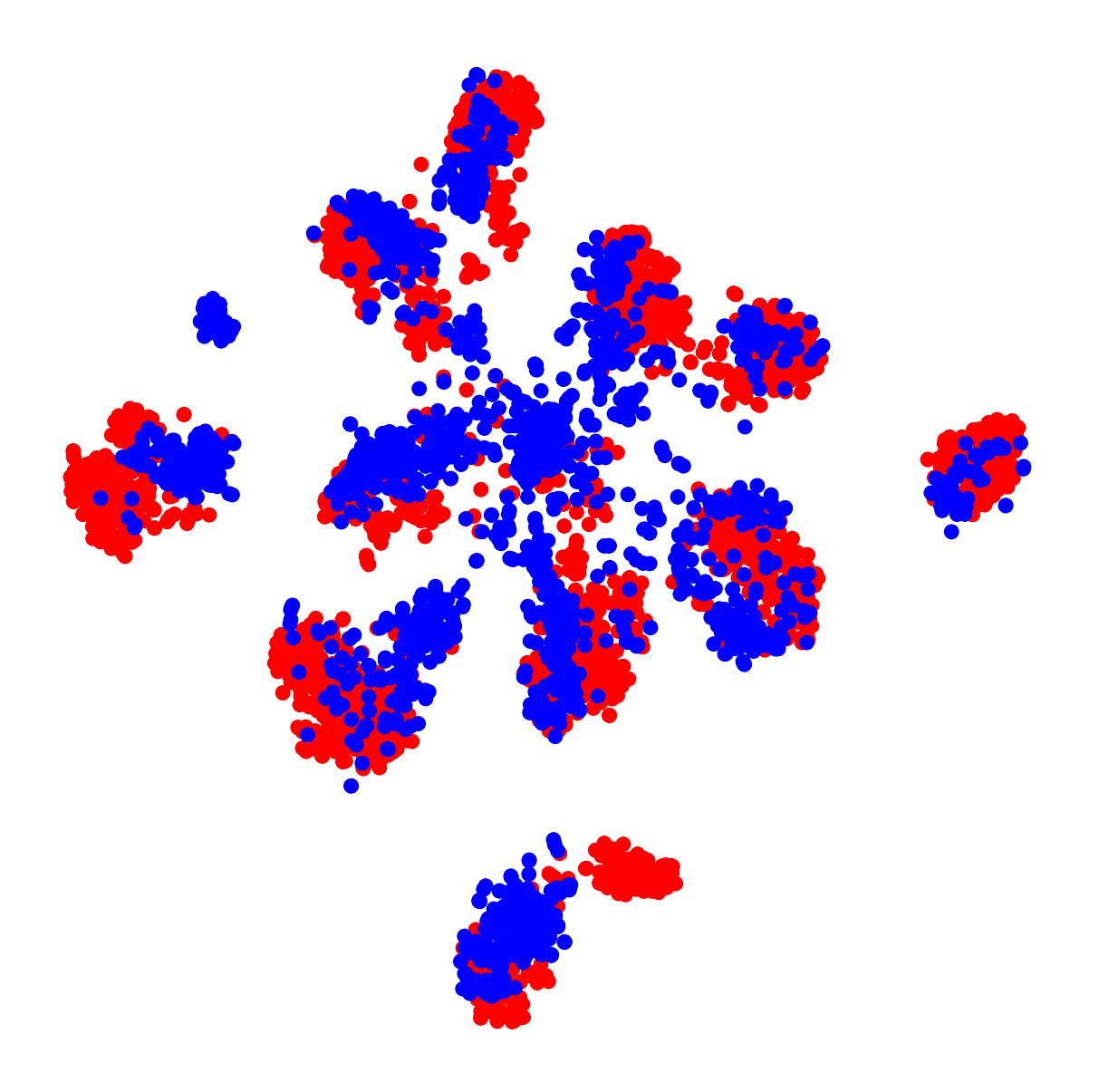}
    }
    \subfigure[]{
    \centering
    \includegraphics[width=1.2in,height=0.9in]{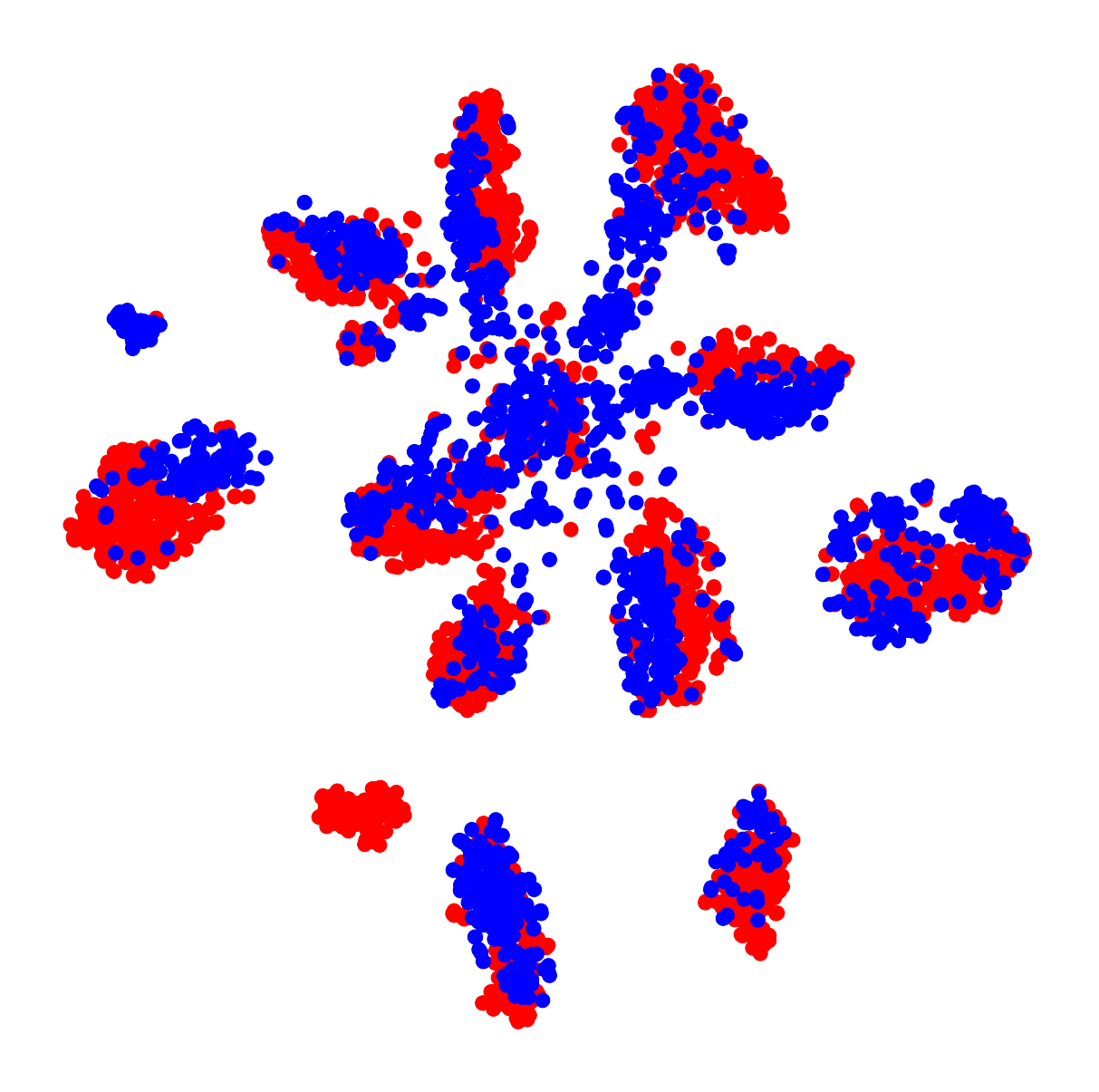}
    }
    \subfigure[]{
    \centering
    \includegraphics[width=1.2in,height=0.9in]{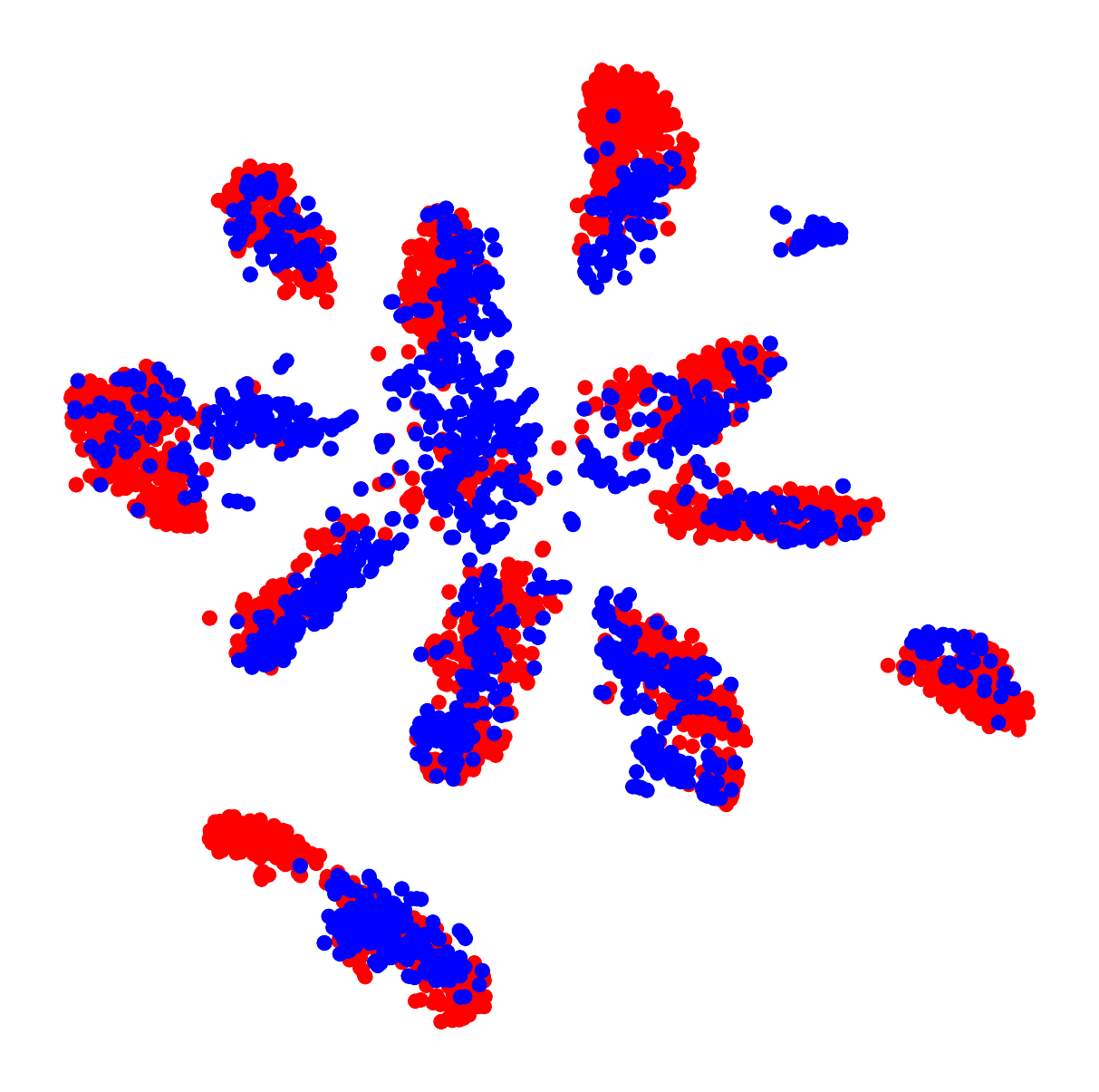}
    }
    \subfigure[]{
    \centering
    \includegraphics[width=1.2in,height=0.9in]{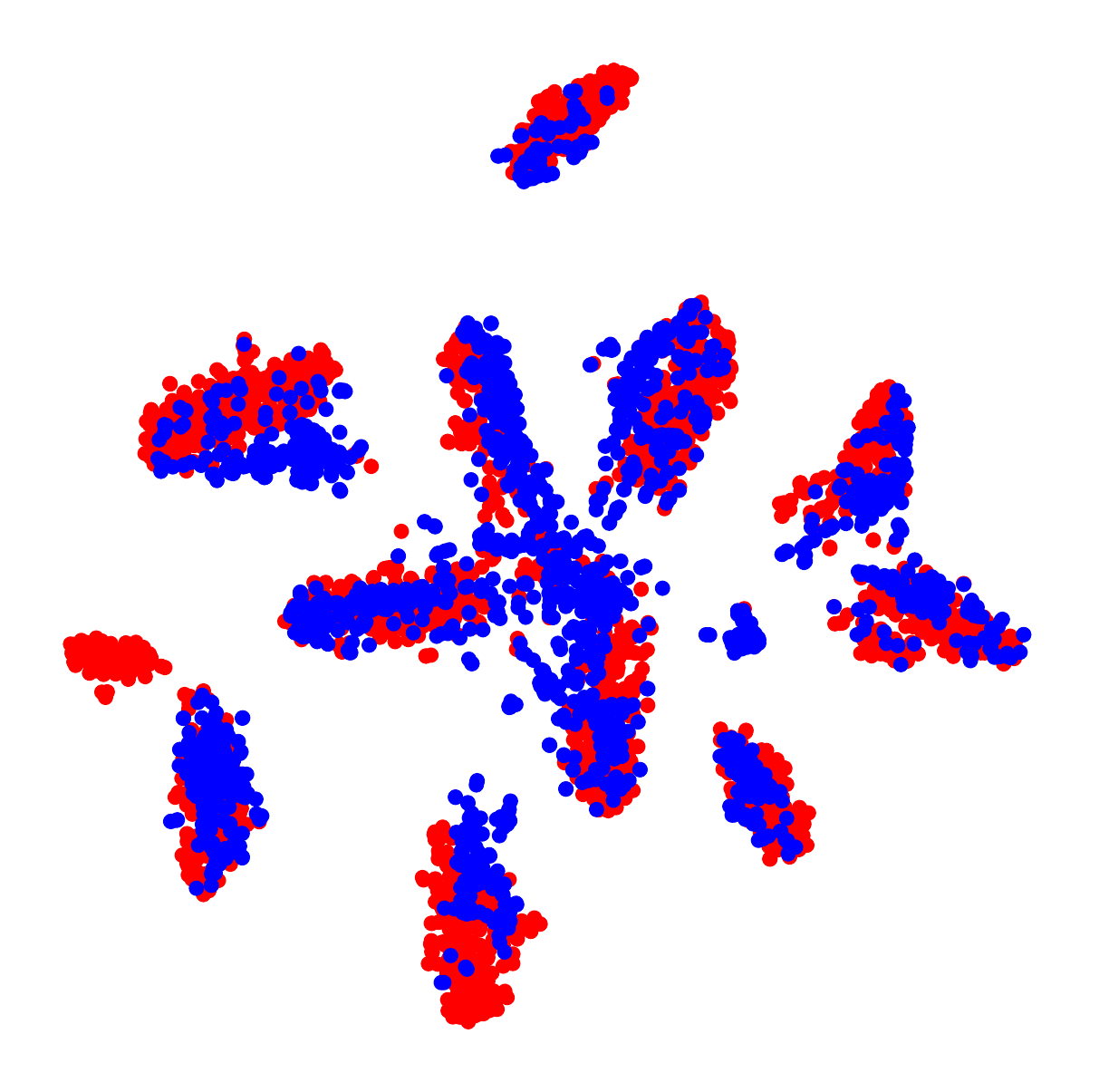}
    }
    
    \subfigure[]{
     \includegraphics[width=1.2in,height=0.9in]{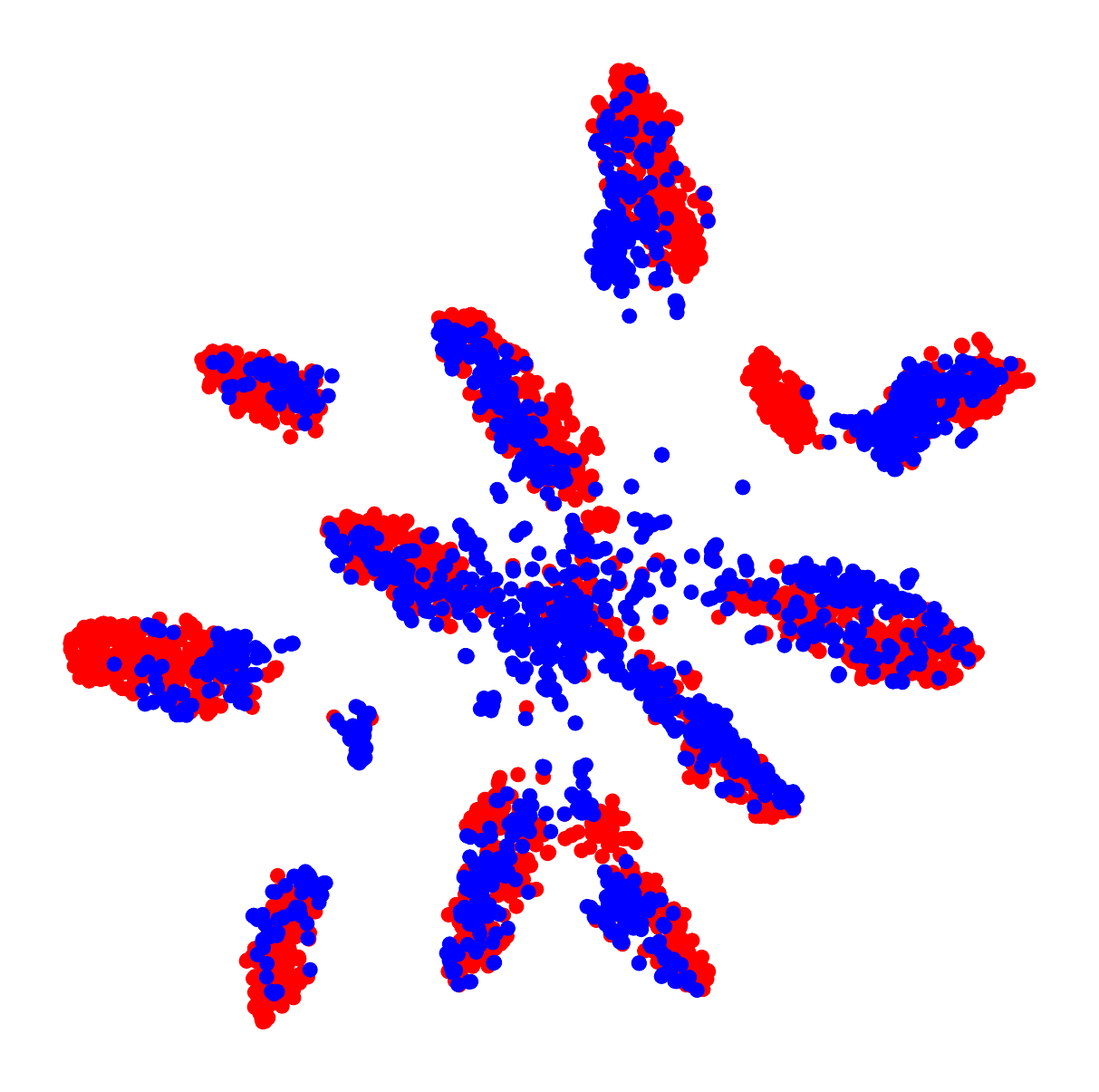}
    }
    \subfigure[]{
     \includegraphics[width=1.2in,height=0.9in]{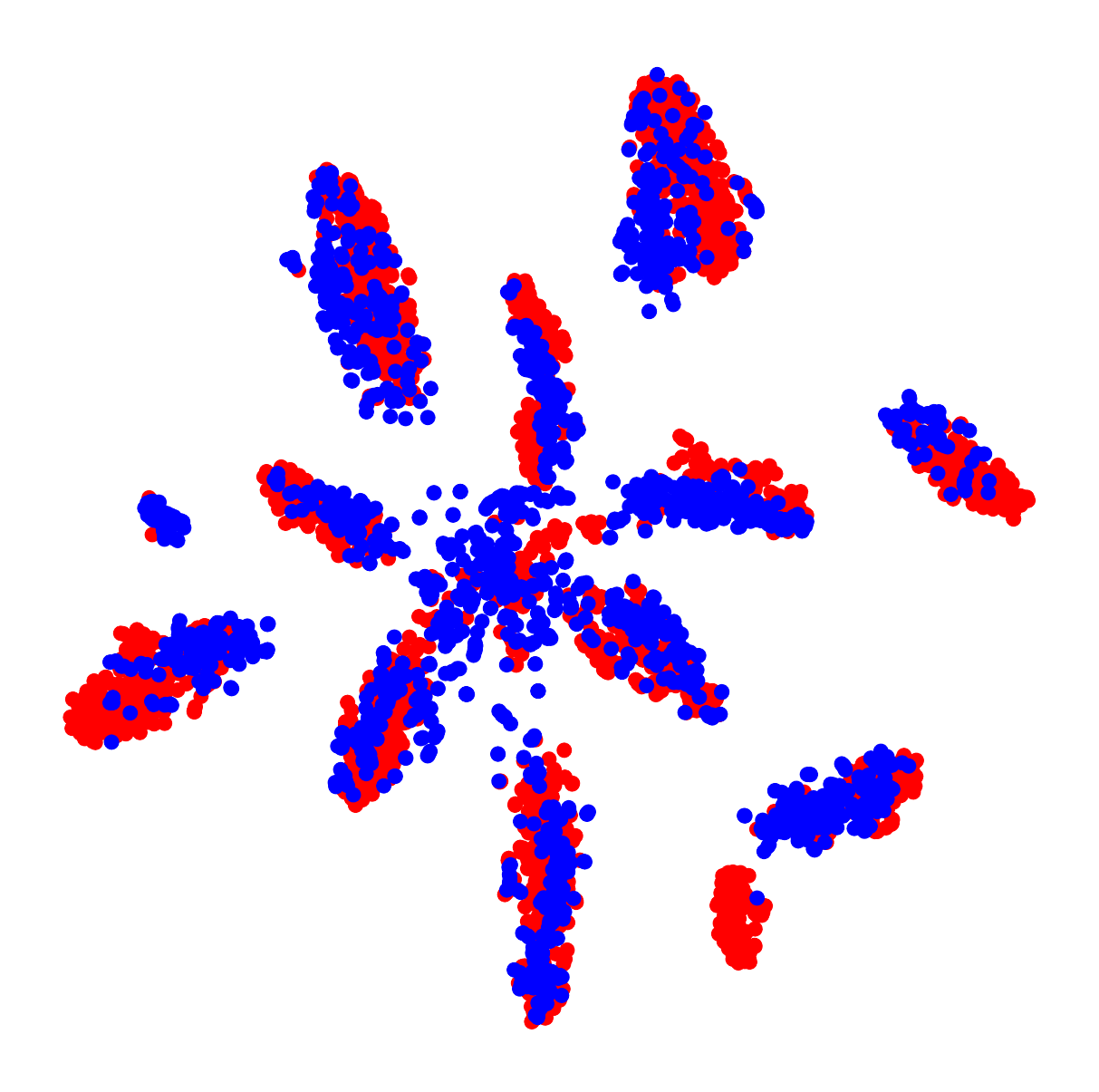}
    }
    \subfigure[]{
     \includegraphics[width=1.2in,height=0.9in]{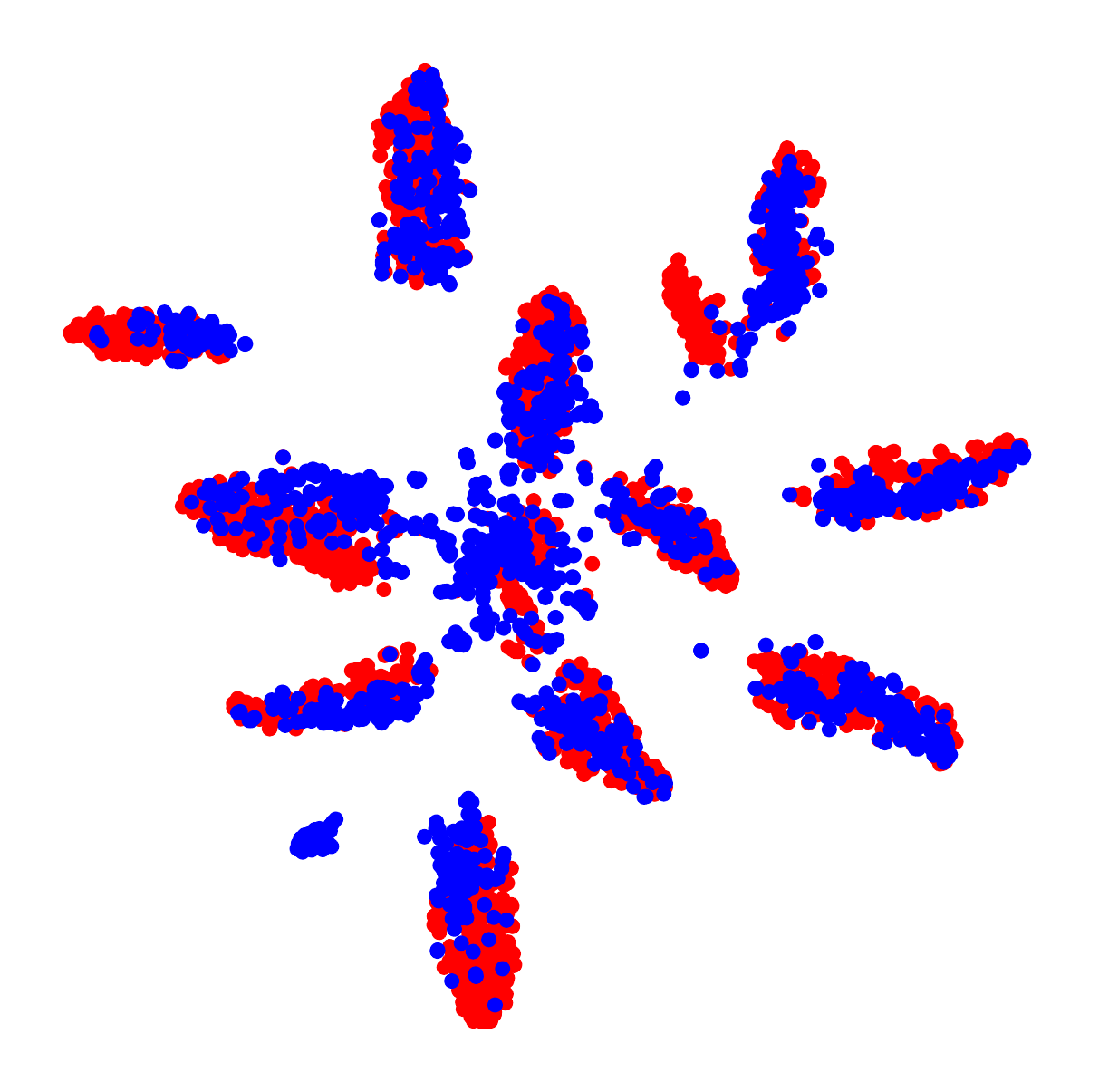}
    }
    \subfigure[]{
     \includegraphics[width=1.2in,height=0.9in]{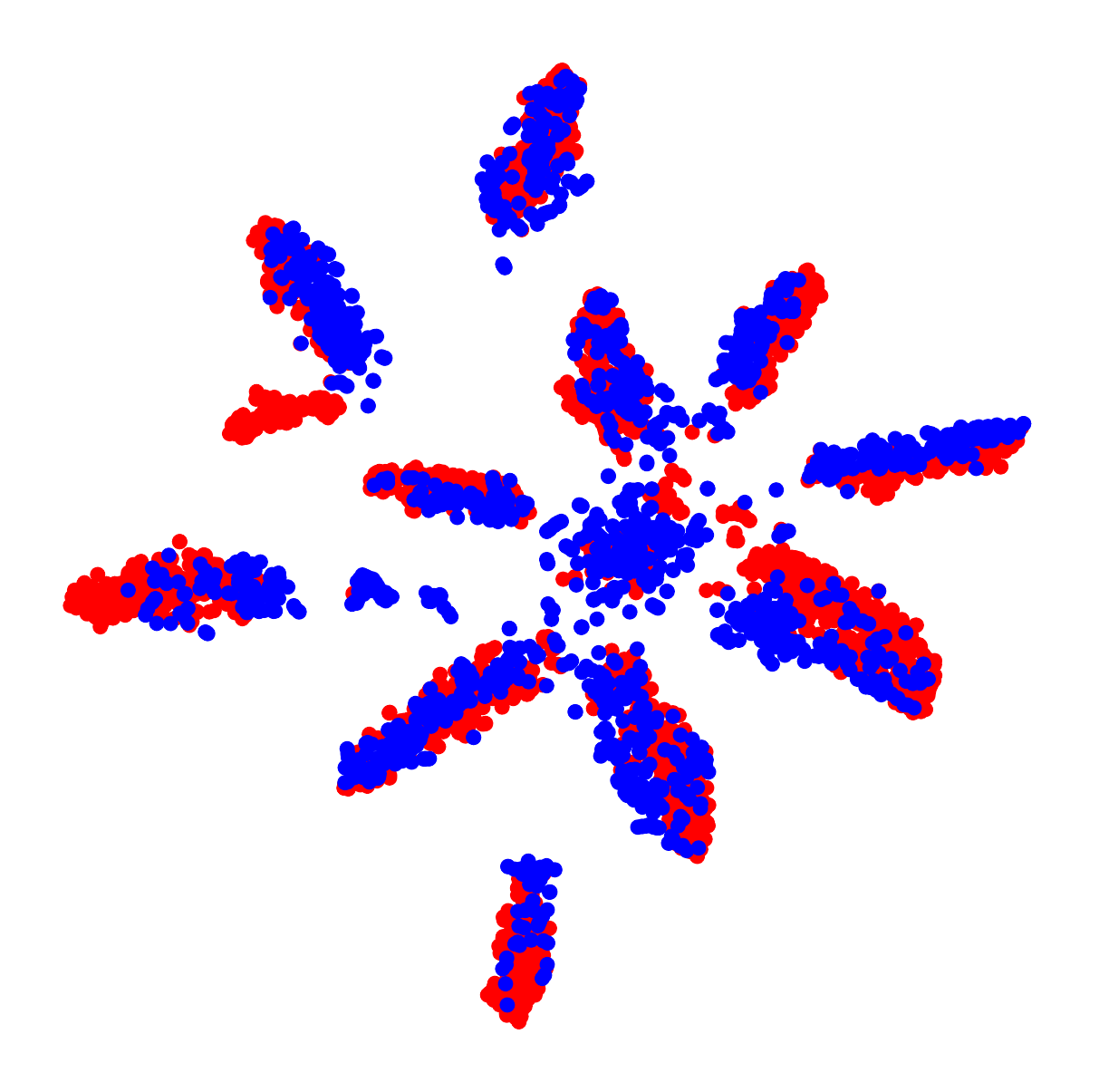}
    }
    \subfigure[]{
     \includegraphics[width=1.2in,height=0.9in]{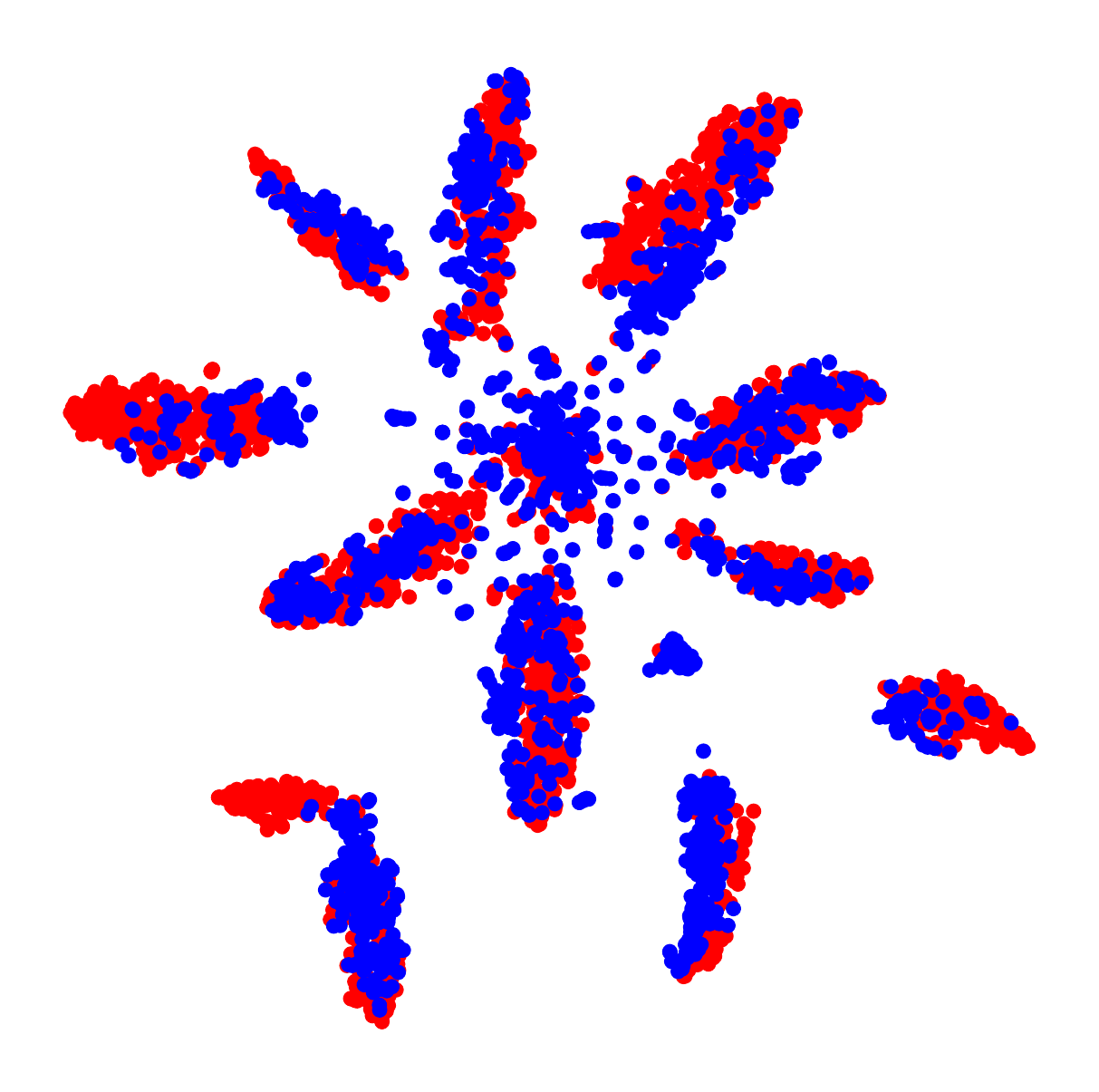}
    }
\caption{The t-SNE~\cite{maaten2008visualizing} visualization results of shared ten-class features in the 3-shot \textbf{R$\rightarrow$S} problem  obtained by our proposed methods (\ie, \bm{$H_{src}$}+\bm{$H_{tar}$}) at the: (a) 1000-th, (b) 2000-th, (c) 3000-th, (d) 4000-th, (e) 5000-th, (f) 6000-th, (g) 7000-th, (h) 8000-th, (i) 9000-th, and (j) 10000-th epoch.}\label{sup:f4}
\end{figure*}

To demonstrate the effectiveness of source entropy loss, e.g., \bm{$H_{src}$}, we visualize the generator features of \bm{$H_{src}$} + \bm{$H_{tar}$} and \bm{$H_{src}$} in Fig.~\ref{sup:f4} and Fig.~\ref{sup:f5} respectively as the main paper. The visualized results are obtained by the generator model trained at 1000-th to 10000-th epoch. According to such figures, it is easy to observe that the cross-domain features become gathering and aligned through the training. With the source entropy loss, e.g., \bm{$H_{src}$}, the features in Fig.~\ref{sup:f4} are more clustered and separable compared with those in Fig.~\ref{sup:f5}. Moreover, \bm{$H_{src}$} + \bm{$H_{tar}$}' features are earlier clustered which makes it fastly converge.

\section{ Unsupervised Domain Adaptation (UDA)}
Although our proposed model is not specifically designed for the unsupervised domain adaptation (UDA), we do believe UODA could be extended to the UDA setting. We slightly modify our model and reveal its potential performance under the UDA setting. The details are introduced below.

\subsection{Objective Functions}
By incorporating source features scattering and target features clustering, the margins between the decision boundary and target features would be enlarged, which enforces the cross-domain feature alignment in a group-by-group way. In this procedure, the labels in the target features provide crucial information to guide the feature alignment. However, the clusters of the target features could preserve some distribution knowledge and we assume that the Opposite Structure Learning strategy still works without the label guidance. To this end, we proposed the UDA version of our method. Due to the lack of $\mathcal{L}_{tar}$, the objective functions for $\mathcal{G}(\cdot)$, $\mathcal{F}(\cdot)_{1}$ and $\mathcal{F}(\cdot)_{2}$ are revised as:

\begin{align}
    {\Theta}_{\mathcal{F}_{1}}^{*} =&\mathop {\arg \min} \limits_{{\Theta}_{\mathcal{F}_{1}}} \mathcal{L}_{src} + \beta H_{src},\label{F_1}\\
    {\Theta}_{\mathcal{F}_{2}}^{*} =&\mathop {\arg \min} \limits_{{\Theta}_{\mathcal{F}_{2}}} \mathcal{L}_{src}  - \lambda H_{tar},\label{F_2}\\
    {\Theta}_\mathcal{G}^{*} =&\mathop {\arg \min} \limits_{{\Theta}_\mathcal{G}} \mathcal{L}_{src} - \beta H_{src} + \lambda H_{tar},
\end{align}
where all the symbols follow the meanings in the main paper.


\subsection{Experiments}
To demonstrate the superiority of UODA under the UDA setting, we comprehensively evaluate it as well as other baseline methods with only the labeled source samples and unlabeled target samples. Details of experiments are described below.


\subsubsection{Experiments Setup}
Following the protocol in MME~\cite{saito2019semi}, we take the AlexNet~\cite{simonyan2014very} as the backbone of the generator $\mathcal{G}(\cdot)$. The assignments of other hyperparamters follow those of the main paper. We take the DomainNet\footnote{\url{http://ai.bu.edu/M3SDA/}}~\cite{peng2019moment} given no labeled target domain samples as the benchmark for the evaluation. \textbf{Since our method is not designed for the UDA problem, we simply compare UODA with other SSDA methods for the fair comparison}.

\subsubsection{Results Analysis}
All the quantitative results on the setting of unsupervised domain adaptation are summarised in Table~\ref{t1}. It's easily observed that ours achieves the best on most of adaptation scenarios as well as on the average result which demonstrate the effectiveness of our method. While on \textit{Painting} to \textit{Clipart} and \textit{Clipart} to \textit{Sketch}, MME~\cite{saito2019semi} slightly outperforms ours which means that source feature scattering has the possibility to introduce negative transfer. In this case, the target features are falsely enclosed by the source features due to the lack of constraints (\textit{i.e.,} labeled target samples) to regulate the feature space.

\begin{table*}[t]
\begin{center}
\caption{Quantitative results $\%$ on DomainNet under the UDA setting.}
\label{t1}
\scalebox{1.0}{
\begin{threeparttable}
 \centering
  \begin{tabular}{|c|ccccccc|c|}
   \hline \hline

   {Methods} & \multicolumn{1}{c}{R$\rightarrow$C} & \multicolumn{1}{c}{R$\rightarrow$P} & \multicolumn{1}{c}{P$\rightarrow$C} & \multicolumn{1}{c}{C$\rightarrow$S} & \multicolumn{1}{c}{S$\rightarrow$P} & \multicolumn{1}{c}{R$\rightarrow$S} & \multicolumn{1}{c}{P$\rightarrow$R} & \multicolumn{1}{|c|}{Avg}\\
   \hline
   
\multicolumn{1}{|c|}{Source Only}   &41.1 &42.6  &37.4 &30.6 &30.0 &26.3 &52.3 &\multicolumn{1}{c|}{37.2}  \\



 
  \multicolumn{1}{|c|}{ENT~\cite{grandvalet2005semi}}  &33.8 &43.0  &23.0 &22.9 &13.9 &12.0 &51.2 &\multicolumn{1}{c|}{28.5} \\
 
\multicolumn{1}{|c|}{MME~\cite{saito2019semi}}  &47.6 &44.7  &\textbf{39.9} &\textbf{34.0} &33.0 &29.0 &53.5 &\multicolumn{1}{c|}{40.2} \\

\multicolumn{1}{|c|}{Ours}  &\textbf{48.4} &\textbf{48.2}  &{39.6} &{33.0} &{34.4} &\textbf{30.6}  &\textbf{58.3}&\multicolumn{1}{c|}{\textbf{41.8}} \\


 \hline \hline
\end{tabular}
\end{threeparttable}
}
\end{center}
\end{table*}

\subsection{Further Discussion}

As it is inaccessible to the labeled target samples under the UDA setting, there is no $\mathcal{L}_{tar}$ as one of the objectivenesses. In the UDA case, aligning the cross-domain features becomes a more important issue compared with those of SSDA which focus on the optimal structure learning and the overcome of overfitting on the target domain. We have not explicitly incorporated the alignment loss in our framework as UODA is designed for the SSDA problem where the cross-domain features would be roughly aligned with the help of partially labeled target samples. Therefore, the best way to extend UODA to the UDA setting is to incorporate feature alignment loss such as GAN loss. In this section, we simply compare the UODA with other SSDA methods following the protocol of MME~\cite{saito2019semi}. Extending UODA to a general DA framework to tackle both SSDA and UDA problems is our next plan.

\end{document}